\pdfoutput=1

\documentclass[11pt]{article}

\usepackage[final]{coling}

\usepackage{times}
\usepackage{latexsym}

\usepackage[T1]{fontenc}

\usepackage[utf8]{inputenc}

\usepackage{microtype}

\usepackage{inconsolata}

\usepackage{graphicx}

\usepackage{amsmath}
\usepackage{annotate-equations}
\usepackage{tikz}
\usepackage[compact]{titlesec}
\usepackage{booktabs}       
\usepackage{caption}
\usepackage{tabularx}
\usepackage{subcaption}
\usepackage{mathtools}
\usepackage{multirow}

\title{DEPTH: Discourse Education through Pre-Training Hierarchically}

\author{%
  Zachary Elisha Bamberger
    \\
  Technion \\
  \texttt{zachary@campus.technion.ac.il} \\
  \And
  Ofek Glick \\
  Technion \\
  \texttt{ofek.glick@campus.technion.ac.il } \\
  \AND
  Chaim Baskin \\
  Ben-Gurion University of the Negev \\
  \texttt{chaimbaskin@bgu.ac.il } \\
  \And
  Yonatan Belinkov \\
  Technion \\
  \texttt{belinkov@technion.ac.il } \\
}

\begin{document}
\maketitle
\begin{abstract}
Language Models (LMs) struggle with linguistic understanding at the discourse level, even though discourse patterns such as coherence, cohesion, and narrative flow are prevalent in their pre-training data. To improve the discourse capabilities of LMs already at the pre-training stage, we introduce DEPTH, an encoder-decoder model that learns latent representations for sentences using a discourse-oriented pre-training objective. DEPTH combines hierarchical sentence representations with two objectives: (1) \textit{Sentence Un-Shuffling}, and (2) \textit{Span-Corruption}. Our approach trains the model to represent both sub-word-level and sentence-level dependencies over a pre-training corpora. When trained either from scratch or continuing from a pre-trained T5 checkpoint, DEPTH learns semantic and discourse-level representations faster than T5, outperforming it in span-corruption loss despite the additional sentence-un-shuffling objective. Evaluations on the GLUE, DiscoEval, and NI benchmarks demonstrate DEPTH's ability to quickly learn diverse downstream tasks, which require syntactic, semantic, and discourse capabilities. Our approach extends the discourse capabilities of T5, while minimally impacting other natural language understanding (NLU) capabilities in the resulting LM. We share ur codebse  for reproducibility: \href{https://github.com/zbambergerNLP/depth.git}{https://github.com/zbambergerNLP/depth.git}.
\end{abstract}

\section{Introduction}

Discourse understanding---the ability to understand how sentences and broader textual units form cohesive narratives \cite{miltsakaki-etal-2004-penn, prasad-etal-2008-penn, jernite2017discoursebasedobjectivesfastunsupervised, prasad-etal-2018-discourse}---is fundamental to effective communication. However, LMs often struggle with this, especially when dealing with long and complex inputs, hindering their performance on tasks like persuasive argumentation \citep{Durmus2019-pragmatic-discourse, hidey-etal-2017-analyzing, chakrabarty-etal-2019-ampersand}, summarization \citep{Zhao2022ReadTN_summarization}, essay scoring \citep{corruption-is-not-all-bad-2021}, dialogue systems \citep{hua-etal-2023-improving-dialogue-summarization}, and following instructions \citep{wei2022finetuned}. Recent evidence from \citet{yu_etal_predicting_the_next_sentence} reinforces this view, demonstrating that human language comprehension occurs at multiple levels and that incorporating discourse-level objectives like next sentence prediction (NSP) can lead to more human-like language representations and improved contextual understanding.

Early attempts to incorporate discourse awareness into pre-training, such as Next Sentence Prediction (NSP) in BERT \citep{devlin-etal-2019-bert} and Sentence Order Prediction (SOP) in ALBERT \citep{Lan2020ALBERT:}, proved overly simplistic and hindered learning effective discourse representations \citep{liu2019roberta, Lan2020ALBERT:, raffel2020exploring_t5}. Subsequent encoder models like Sentence-level Language Model (SLM) \citep{lee-etal-2020-slm}, CONPONO \citep{iter-etal-2020-pretraining-compono}, and Hi-Transformer \citep{wu-etal-2021-hi} improved discourse capabilities in LMs but lacked generative capabilities. 

Unlike the pre-training tasks for encoder LMs, next-token prediction provides decoder LMs like GPT \citep{radford2018improving, radford2019language, gpt3, openai2023gpt4} with powerful generative capabilities. 
However, without a dedicated and costly alignment phase \citep{rlhf_paper_ouyang, wei2022finetuned, bai2022constitutional}, these LMs tend to falter
in understanding and executing human queries.

Even with a dedicated alignment phase, large decoder-only models generally perform poorly on discourse-oriented benchmarks that measure coherence and cohesion \citep{chen-etal-2019-evaluation-discoeval, maimon-tsarfaty-2023-cohesentia, maimon-2023-coherence-assessment, wang2023discobenchdiscourseawareevaluationbenchmark}. Encoder-decoder models such as T5 \citep{raffel2020exploring_t5} and BART \citep{lewis-etal-2020-bart} consistently outperform much larger ($\approx400\times$) decoder-only models like GPT-3 \citep{gpt3} and GPT-4 \cite{openai2023gpt4} on these tasks. Recent work by \citet{katz2024segmentbasedattentionmaskinggpts} indicates that incorporating encoder-decoder attention mechanisms into modern decoder-only models like Llama-3 \cite{grattafiori2024llama}, Qwen-2.5 \cite{qwen2025qwen25technicalreport}, and Mistral \cite{jiang2023mistral7b} also boosts performance, suggesting potential additional benefits from pre-training models with such attention schemes.

To improve the discourse-capabilities of encoder-decoders already at the pre-training stage, we propose DEPTH (\textbf{D}iscourse-\textbf{E}ducation through \textbf{P}re-\textbf{T}raining \textbf{H}ierarchically), a hierarchical language model that learns representations for both sub-word and sentence-level tokens. 
DEPTH extends the pre-training objective of SLM from encoder-only models like BERT, to encoder-decoder models like T5.
Notably, DEPTH introduces latent, heirarchical representations for sentences (as in \citet{lee-etal-2020-slm, yang-etal-2021-universal, yu_etal_predicting_the_next_sentence}) directly into the objective of a \textit{generative} pre-training task.
By employing a hierarchical attention mechanism across sub-word and sentence level tokens, DEPTH captures both fine-grained semantic dependencies and broader inter-sentence relationships. 
When pre-trained from scratch, our DEPTH model obtains meaningful representations for downstream tasks much faster than our baseline T5. Continuously pre-training T5 models with the DEPTH objective improves the discourse capabilities of the resulting models, without sacrificing performance in downstream NLU tasks. 

\section{Method}
\label{sec:method}

Pre-training DEPTH involves simultaneously performing span corruption \citep{raffel2020exploring_t5}, while also un-shuffling sentences as in \citet{lee-etal-2020-slm}. In Section \ref{sec:tokenization} we introduce a new tokenizer for DEPTH, which combines T5's tokenizer with the sentence segmentation operation required for sentence un-shuffling. In Section \ref{sec:corruption}, we detail how we combined the pre-training objectives of both models into the sequence-to-sequence framework of T5. Next, in Section \ref{sec:attention_masks}, we discuss how we introduce hierarchical representations during pre-training, and how this hierarchy encourages the model to learn discourse representations. Finally, in Section \ref{sec:loss_formulation}, we demonstrate how to combine the losses of T5 and SLM into a unified objective that is conducive to traditional teacher-forcing.

\subsection{Tokenization}
\label{sec:tokenization}

We introduce a tokenization function, $t$, which transforms an input string, $s$, into a sequence of tokens, $X=(x_{1,1}, x_{1,2}, \ldots, x_{m, \text{len}(m)})$, in our vocabulary, $V$. Each token $x_{i, j}$ denotes the $j$'th token of the $i$'th sentence, where $m$ is the number of sentences and $\text{len}(i)$ is the length of the $i$'th sentence. $V$ includes special tokens \texttt{<EOS>}, \texttt{<BOS>}, \texttt{<PAD>}, and sentinel tokens $V_{\text{sentinel}}$ of the form \texttt{<special\_token\_z>} as in the original T5 paper.

\begin{figure*}[h!]
    \centering
    \includegraphics[width=0.7\textwidth]{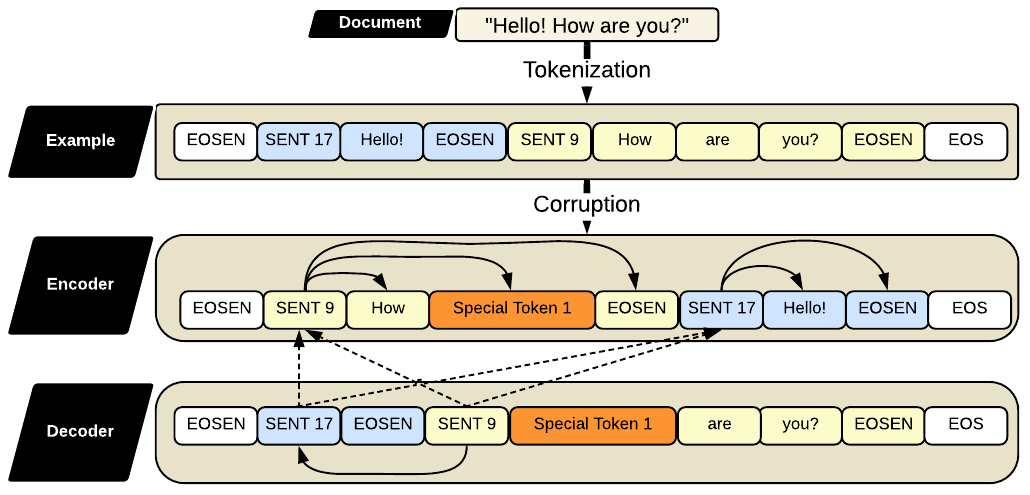}
    \caption{DEPTH tokenization and corruption process. Given an input document, DEPTH introduces sentence tokens (\texttt{<SENT\_i>} and \texttt{<EOSEN>}), applies span masking, and shuffles sentences with probability 0.5. Attention patterns are shown with arrows (dotted for cross-attention, solid for self-attention).}
    \label{fig:depth_pipeline}
\end{figure*}

To facilitate DEPTH's hierarchical structure, we segment sentences using NLTK \citep{bird-loper-2004-nltk} and create $k+1$ new tokens\footnote{We follow \citet{lee-etal-2020-slm}, using $k=20$ sentence tokens.}:
\begin{align*}
S  &= \{ \texttt{<SENT\_1>}, \ldots, \texttt{<SENT\_k>}, \texttt{<EOSEN>} \} \\
V' &= V \cup S
\end{align*}

We augment our tokenizer function $t$ to form $t'$, which maps sequence $s$ to tokens in $V'$. In each sentence, we prepend a \texttt{<SENT\_i>} token and append a \texttt{<EOSEN>} token:
\begin{align*}
X = \{ &\texttt{<SENT\_a>}, x_{1, 1}, x_{1, 2}, \ldots, x_{1, \text{len}(1)}, \texttt{<EOSEN>}, \\
       &\ldots, x_{m, \text{len}(m)}, \texttt{<EOSEN>}, \texttt{<EOS>} \}
\end{align*}

The integer $i$ in \texttt{<SENT\_i>} represents a sentence's index, sampled uniformly at random from $\{1, \ldots, k\}$ without replacement. We truncate sentences beyond the $k$'th to limit vocabulary size.

Unlike SLM (an encoder-only LM with an auxiliary pointer-decoder), DEPTH is an encoder-decoder that predicts \texttt{<SENT\_i>} and \texttt{<EOSEN>} token IDs directly. The \texttt{<EOSEN>} token signals the next token is either \texttt{<SENT\_i>} or \texttt{<EOS>}, allowing for dynamic attention masking in the decoder.

Formally, we define a pre-tokenization function $f: s \rightarrow s'$, where $s'$ includes \texttt{<SENT\_i>} and \texttt{<EOSEN>} tokens. The tokenized input for DEPTH is produced with $T(s) = t(f(s)) = X$. We use the SentencePiece \citep{kudo-richardson-2018-sentencepiece} tokenizer as $t$, adjusted to support DEPTH's sentence-level pre-training objective.

\subsection{Corruption}
\label{sec:corruption}

\paragraph{Span-Masking:} We apply a corruption process to each batch of tokenized sequences. We sample masked spans using a geometric distribution (as in \citet{joshi-etal-2020-spanbert} and \citet{ raffel2020exploring_t5}), parameterized with an average span length of $\lambda$ and a masking probability of $p$. Spans overlapping with sentence tokens (\texttt{<EOSEN>} or \texttt{<SENT\_i>}) are ignored. Masked token spans are replaced with a single sentinel token \texttt{<special\_token\_z>}, where ($z$) is a uniformly sampled integer from ${0, \ldots, 99}$. The missing token sequences appear after the corresponding sentinel token in the labels. Note that in the original T5 implementation, sentinel tokens appear in the same incrementally decreasing order in each example (\texttt{<special\_token\_99>} followed by \texttt{<special\_token\_98>}, etc...). We randomly sample sentinel token ID's for DEPTH to eliminate hints about sentence positions from the sentinel tokens. For example, with the T5 scheme for span-masking, the presence of \texttt{<special\_token\_99>} indicates that the sentence in which it appears comes first, making sentence un-shuffling too easy.

\paragraph{Sentence Un-Shuffling:}
Given an input sequence of up to $k$ sentences, we randomly shuffle the order of sentences in the model input\footnote{We do not shuffle the order of tokens within a sentence}. We then task the model with reconstructing the original order of the sentence tokens in the target. We shuffle all examples in a batch with probability $p=0.5$ (as in SLM). 
By disrupting the original sentence order, DEPTH encourages learning of (1) the complete meaning of individual sentences, independent of their surrounding context, and (2) representations that encode how sentences relate to each other\footnote{E.g., discourse markers, co-reference, and entailment}. We show DEPTH's corruption process in Figure \ref{fig:depth_pipeline}.

\subsection{Attention masks}
\label{sec:attention_masks}
Our baseline model (T5) utilizes bidirectional attention in the encoder, auto-regressive attention in the decoder, and full cross attention from the decoder to the encoder. However, T5's formulation does not account for the hierarchical treatment of sentences used by SLM and DEPTH.

We define \textit{non-sentence} tokens, $x_{reg}$, as tokens $x \in X$ where $x \notin S$. We compose attention masks to impose hierarchy. As part of encoder self-attention, non-sentence tokens can attend to all other tokens in the corrupted input sequence, while \texttt{<SENT\_i>} tokens can only attend to tokens within their own sentence (including sentinel tokens). As part of decoder self-attention, all tokens have an auto-regressive attention mask, but \texttt{<SENT\_i>} tokens can only attend to past sentence tokens. Finally, as part of cross attention, non-sentence tokens in the decoder can attend to the entire input in the encoder, while sentence tokens in the decoder can only attend to sentence tokens in the encoder.  This scheme is depicted in Figure \ref{fig:depth_pipeline}.

This scheme encourages the model to use sentence token representations in the encoder to predict the next sentence token in the decoder via cross-attention. It also ensures that non-sentence tokens in the encoder provide relevant discourse information to their corresponding sentence tokens.

\subsection{Loss Formulation}
\label{sec:loss_formulation}

Let \( Y= \{ y_{1, 1}, y_{1, 2}, \ldots, y_{m, \text{len}(m)} \} \) be the target sequence, where each token \( y_{i,j} \) belongs either to the span-masking task (non-sentence tokens) or to the sentence un-shuffling task (sentence tokens). We denote by \(\hat{S}\) the set of all sentence tokens in \(Y\). The model prediction is given by $
\hat{Y} = \{ \hat{y}_{1, 1}, \hat{y}_{1, 2}, \ldots, \hat{y}_{m, \text{len}(m)} \}$, 
where \(\hat{y}_{i,j}\) represents the predicted probability distribution over the vocabulary for token \( y_{i,j} \). Let the total number of tokens be $N = |Y|$.

The loss for DEPTH, which jointly optimizes the reconstruction (span-masking) and sentence un-shuffling tasks, is defined as the token-averaged cross-entropy:
\vspace{-5pt}
\begin{multline*}
  L_{\text{DEPTH}} = -\frac{1}{N} \sum_{i=1}^{m} \sum_{j=1}^{\text{len}(i)} y_{i,j} \cdot \log \hat{y}_{i,j} \\
  = \eqnmarkbox[blue]{sl}{-\frac{1}{N}\sum_{\mathclap{y_{i,j} \in Y \cap \hat{S}}} y_{i,j} \cdot \log \hat{y}_{i,j}}
  \eqnmarkbox[red]{rl}{-\frac{1}{N}\sum_{\mathclap{y_{i,j} \in Y \setminus \hat{S}}} y_{i,j} \cdot \log \hat{y}_{i,j}}
\end{multline*}
\annotate[yshift=-1em]{below,left}{sl}{Sentence Loss}
\annotate[yshift=-1em]{below,left}{rl}{Reconstruction Loss}
\vspace{20pt}

In this formulation, the summation runs over all sentences in the input (i.e., up to $m$ sentences, where $1\leq m \leq k$), and within each sentence over its tokens. This allows us to decompose the model's performance into the contributions from the sentence un-shuffling and the span-masking tasks. We explore additional loss formulations and weighting schemes in Appendix \ref{sec:sentence_token_weight}.

\section{Experimental setup}
\label{sec:experimental_setup}

The aim of our experiments is to measure the effectiveness of DEPTH against a standard encoder-decoder model. Accordingly, our experiments explore the learning dynamics of DEPTH model relative to a T5-Base (220M parameter) baseline. We pre-train both models on the C4 dataset \citep{raffel2020exploring_t5, dodge-etal-2021-documenting_c4} to resemble the manner in which the original T5 was trained (see additional reasoning in Appendix \ref{sec:pre_training_data}). We chose to use Base-sized models given limited computational resources, and ease of reproducibility (following the example of  \citet{lee-etal-2020-slm, levine2020pmi, zhang2020pegasus, raffel2020exploring_t5}). We did not use SLM as our baseline since its codebase and checkpoints are not released, and it cannot perform free-form text generation.

We chose to run our experiments without example packing since this is how the SLM model, which inspired DEPTH, was trained. Furthermore, example-packing in T5 enables unrelated examples to impact the model's decisions, thereby harming performance \citep{krell2021efficientsequencepacking, shi2024incontext}. While example-packing is critical for more computationally-efficient training \citep{ding2024fewertruncations}, we were interested in measuring which model was able to use training examples more effectively.  
We examine the impacts of avoiding example packing in Appendix \ref{sec:example_packing}.

Consistent with \citet{nawrot-2023-nanot5}, we found that the Adafactor optimizer \citep{adafactor_shazeer_2018}, while more computationally efficient, slightly harmed model performance. We therefore use AdamW \citep{loshchilov2018decoupledAdamW} instead. We use a linearly increasing learning rate during the first $10,000$ steps, and then reduce the learning rate linearly for DEPTH (as in SLM), and with an inverse square learning rate ratio for T5 (as in the original T5 paper). 
We chose to use a masking probability of $p=0.3$,\footnote{\citet{raffel2020exploring_t5} reports that this span corruption ratio does not adversely impact downstream performance, although recently \citet{ankner-etal-2024-dynamic} suggested a dynamic masking rate tends to perform best.} and an average span length of $\lambda=3$. Our mask probability is higher than the advised $0.15$ from T5 to accommodate for the fact that sentence-tokens within DEPTH cannot be masked.  

\noindent We conduct two types of pre-training experiments: 

\begin{enumerate}
    \item \textbf{From Scratch (FS)}: Both T5 and DEPTH models are randomly initialized, and pre-trained on C4 with their respective objectives.

    \item \textbf{Continuous Pre-Training (CPT)}: Both T5 and DEPTH models are initialized from the T5-Base checkpoint on HuggingFace \citep{wolf2019huggingface}, and continue to pre-train on C4 with their respective objectives. 
\end{enumerate}

We note that our CPT experiments build on top of T5 models that have been trained for over 1T tokens, whereas the amount of tokens they see during continuous pre-training is relatively minuscule ($\approx67\times$ fewer tokens for T5, and $\approx80\times$ fewer tokens for DEPTH). We compare configurations of similar-sized models in Appendix \ref{sec:pre_training_hyper_parameters}.

\subsection{Fine-tuning experiments}

We follow up our pre-training experiments with a collection of downstream tasks. We evaluate our models on natural language inference (MNLI, 
\citet{williams-etal-2018-broad}), sentiment analysis (SST2, \citet{socher-etal-2013-recursive}), and grammar (CoLA,
\citet{warstadt2019neural}) within the GLUE benchmark \citep{wang-etal-2018-glue}. 
We also use the DiscoEval suite \citep{chen-etal-2019-evaluation-discoeval} to evaluate models on their understanding of discourse. We use two tasks from DiscoEval: Sentence Permutation (SP) and Discourse Coherence (DC). SP involves identifying the correct position of a removed, while DC involves predicting whether or not a paragraph was coherent. Finally, we measure our model's generative abilities on the Natural Instructions (NI) dataset \citep{mishra-etal-2022-cross}, which measures the ability of LMs to follow instructions, and served as a benchmark for NanoT5 \citep{nawrot-2023-nanot5}.

Our experimental framework is inspired by Pythia \citep{pythia_biderman2023emergent}, which evaluates the performance of LMs on downstream tasks from intermediate checkpoints. We run evaluation with checkpoints from both T5 and DEPTH models, gathered at steps $\{ 2K, 4K, \ldots , 512K, 1M \}$ in order to examine these models' emergent capabilities. The exponential distance between these checkpoints allows us to scale intermediate checkpoint evaluation to much longer training runs.\footnote{Evaluating intermediate checkpoint performance every 10,000 steps (as was done in Pythia) on datasets as large as MNLI is unfeasible with our limited computational resources.}

\section{Results}
\label{sec:results}

\subsection{C4 pre-training}

\begin{figure}[t]
    \centering
    \begin{minipage}{0.8\columnwidth}
        \centering
        \includegraphics[width=\linewidth]{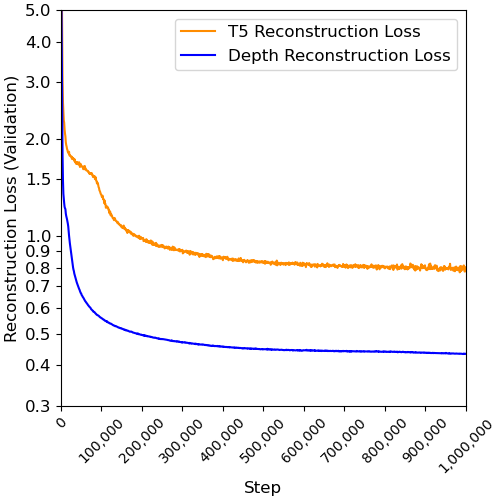}
        \caption{\small From Scratch Pre-Training loss (validation) for both T5 and DEPTH}
        \label{fig:depth_vs_t5_reconstruction_losses}
    \end{minipage}
    \vspace{2mm}
    
    \begin{minipage}{0.8\columnwidth}
        \centering
        \includegraphics[width=\linewidth]{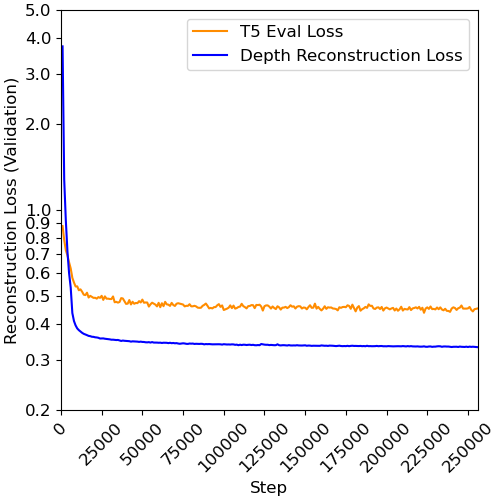}
        \caption{\small Continuous Pre-Training loss (validation) for both T5 and DEPTH.}
        \label{fig:depth_validation_losses}
    \end{minipage}
\end{figure}

During pre-training, we find that DEPTH consistently achieves a lower validation loss than a comparably trained T5 model. This is true for both FS and CPT. Furthermore, when we isolate the reconstruction loss (the objective used by T5, without sentence tokens), we find that DEPTH outperforms T5 despite balancing an additional pre-training objective (Figure \ref{fig:depth_vs_t5_reconstruction_losses} for FS and Figure \ref{fig:depth_validation_losses} for CPT). These results are consistent with the findings in SLM, where their model converged faster, and on fewer tokens than models such as BERT and T5.

While we were not able to match the performance of the baseline model of \citet{raffel2020exploring_t5} (see Appendix \ref{sec:replicating_t5} for speculations on why), we have obtained the lowest loss scores among PyTorch implementations of T5 models. Specifically, in our FS setting, we find that our randomly initialized T5 model outperforms the validation loss of NanoT5, achieving $1.65$ vs.\ $1.95$ at step 64,000.

\subsection{GLUE fine-tuning}

\begin{figure*}[h!]
    \centering
    \begin{subfigure}[b]{0.32\textwidth}
        \includegraphics[width=\textwidth]{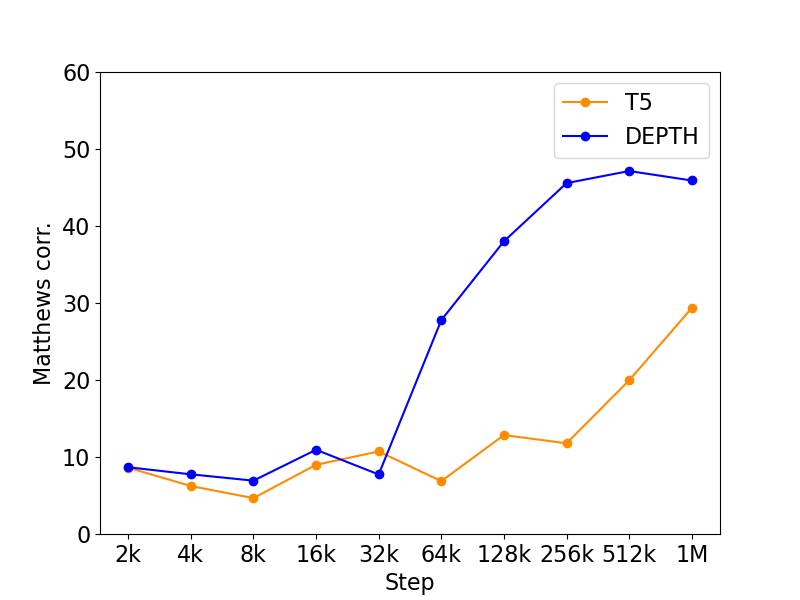}
        \caption{CoLA FS}
        \label{fig:cola_fs}
    \end{subfigure}
    \hfill
    \begin{subfigure}[b]{0.32\textwidth}
        \includegraphics[width=\textwidth]{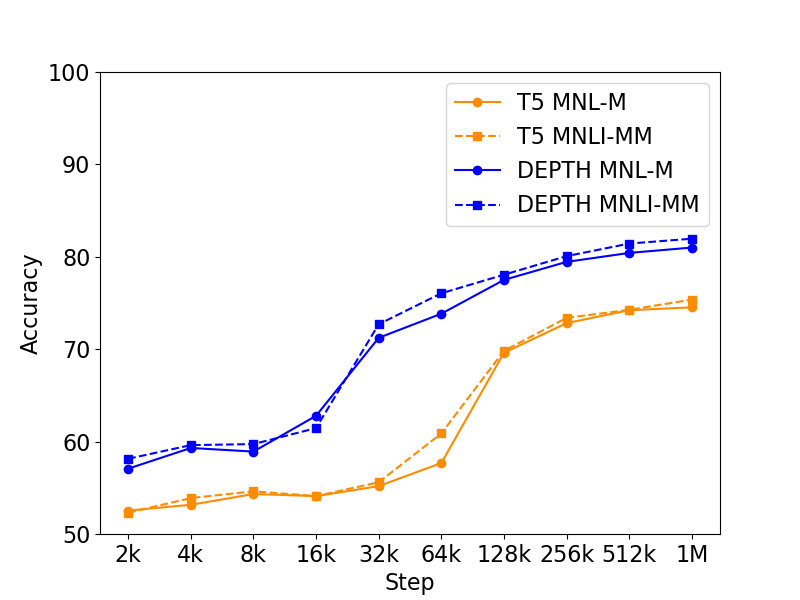}
        \caption{MNLI FS}
        \label{fig:mnli_fs}
    \end{subfigure}
    \hfill
    \begin{subfigure}[b]{0.32\textwidth}
        \includegraphics[width=\textwidth]{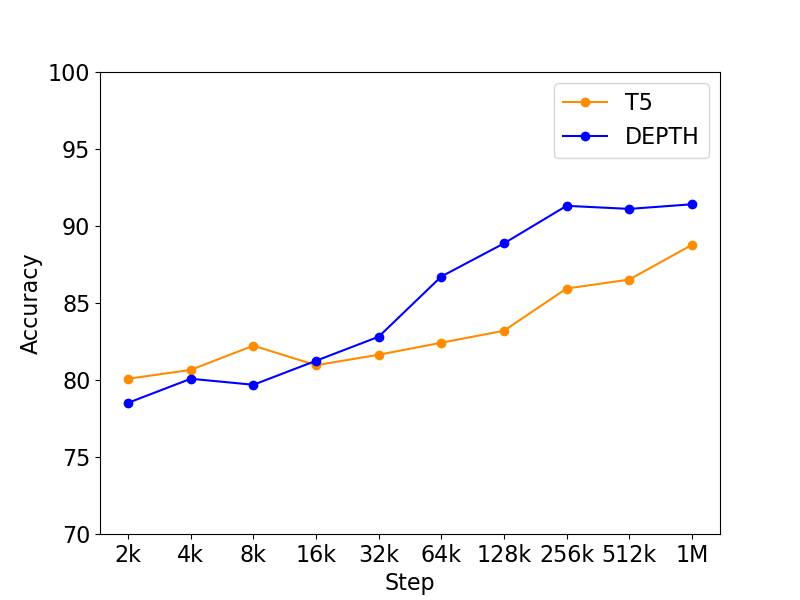}
        \caption{SST-2 FS}
        \label{fig:sst2_fs}
    \end{subfigure}

    \begin{subfigure}[b]{0.32\textwidth}
        \includegraphics[width=\textwidth]{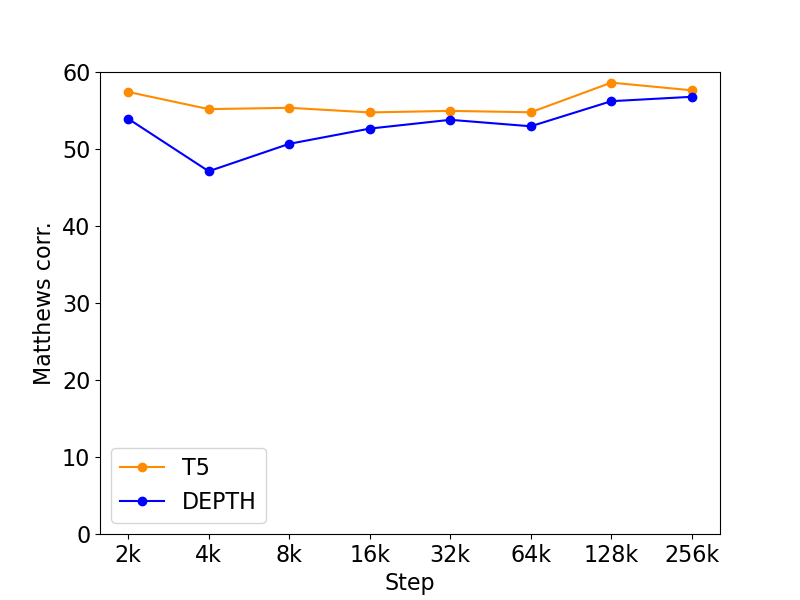}
        \caption{CoLA CPT}
        \label{fig:cola_cpt}
    \end{subfigure}
    \hfill
    \begin{subfigure}[b]{0.32\textwidth}
        \includegraphics[width=\textwidth]{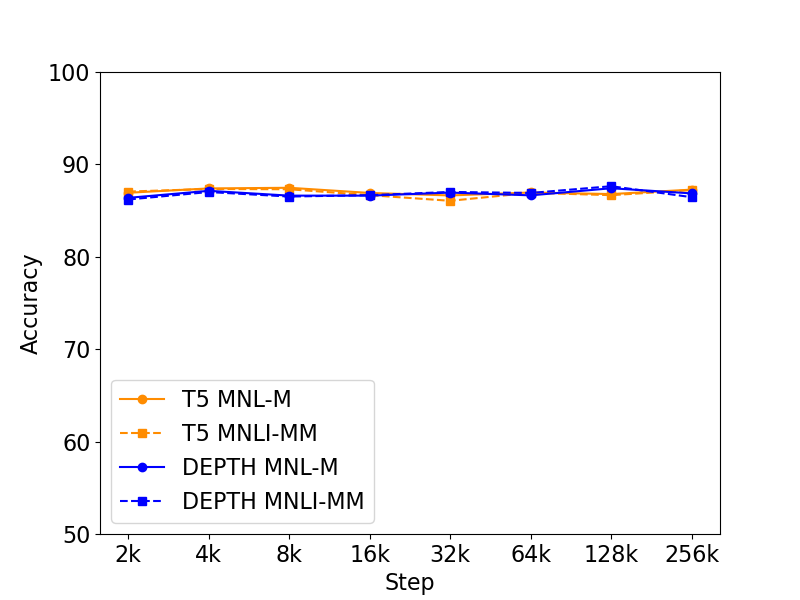}
        \caption{MNLI CPT}
        \label{fig:mnli_cpt}
    \end{subfigure}
    \hfill
    \begin{subfigure}[b]{0.32\textwidth}
        \includegraphics[width=\textwidth]{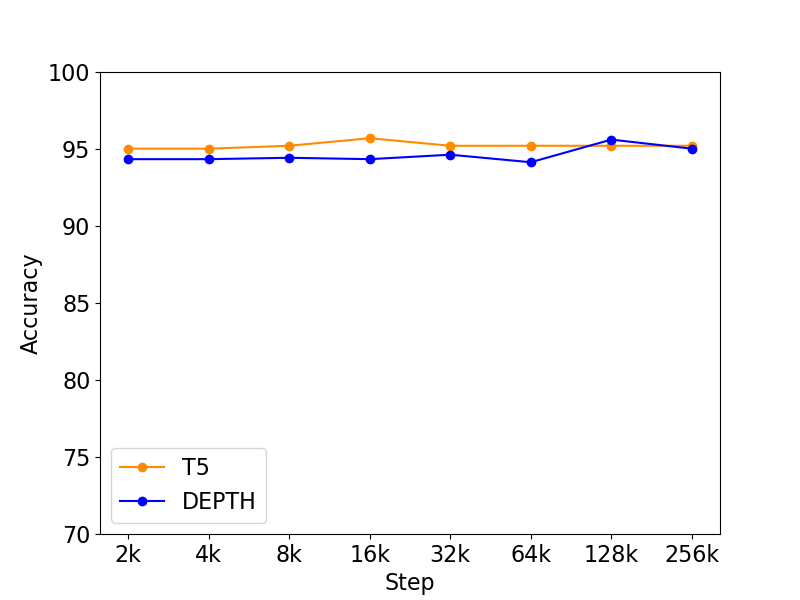}
        \caption{SST-2 CPT}
        \label{fig:sst2_cpt}
    \end{subfigure}

    \caption{GLUE results for FS and CPT models. Top row: From Scratch (FS), Bottom row: From Pretrained (CPT).}
    \label{fig:combined_glue_results}
\end{figure*}

We found that over the course of FS pre-training, both models improved on GLUE tasks. However, T5's improvement pattern was slower than DEPTH's (top row of Figure \ref{fig:combined_glue_results}). We found it difficult to replicate the results of the original T5 (both on GLUE tasks and the pre-training loss) as discussed in Appendix \ref{sec:replicating_t5}. We project that with more substantial training (i.e., 1--3M pre-training steps, and $\geq 2048$ examples per batch, as in T5), DEPTH could match or exceed the performance of T5 and SLM on downstream tasks. 

In the CPT setting (Figure \ref{fig:combined_glue_results}, bottom row), we found that DEPTH and T5 perform similarly, both improving only slightly beyond the baseline. Fine-tuning DEPTH on early CPT checkpoints performs worse than fine-tuning comparable T5-CPT checkpoints. We speculate that this dip in performance is related to the change in objective from span-masking to span-masking \textit{and} sentence un-shuffling. We share our full results on GLUE in Appendix \ref{sec:glue_results_appendix}.

\subsection{DiscoEval fine-tuning}

\begin{figure*}[h!]
    \centering
    \begin{minipage}{0.8\textwidth}
    \begin{subfigure}[b]{0.48\linewidth}
        \includegraphics[width=\linewidth]{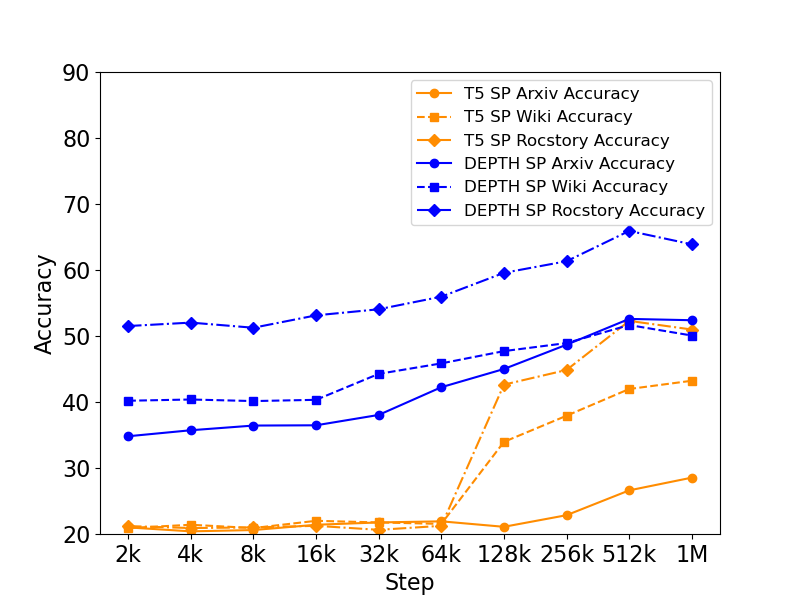}
        \caption{\footnotesize Sentence Permutation (SP) FS}
        \label{fig:sp_fs}
    \end{subfigure}
    \hfill
    \begin{subfigure}[b]{0.48\linewidth}
        \includegraphics[width=\linewidth]{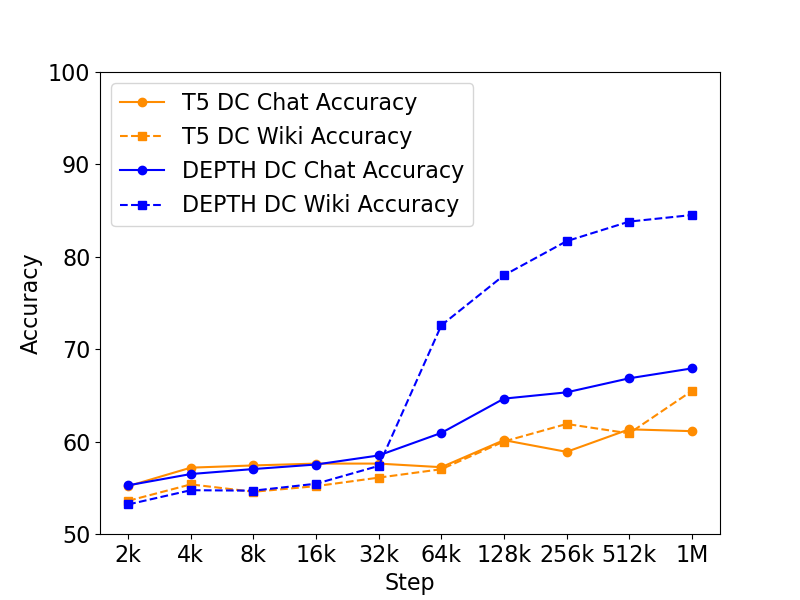}
        \caption{\footnotesize Discourse Coherence (DC) FS}
        \label{fig:dc_fs}
    \end{subfigure}
    
    \vspace{-2ex}
    
    \begin{subfigure}[b]{0.48\linewidth}
        \includegraphics[width=\linewidth]{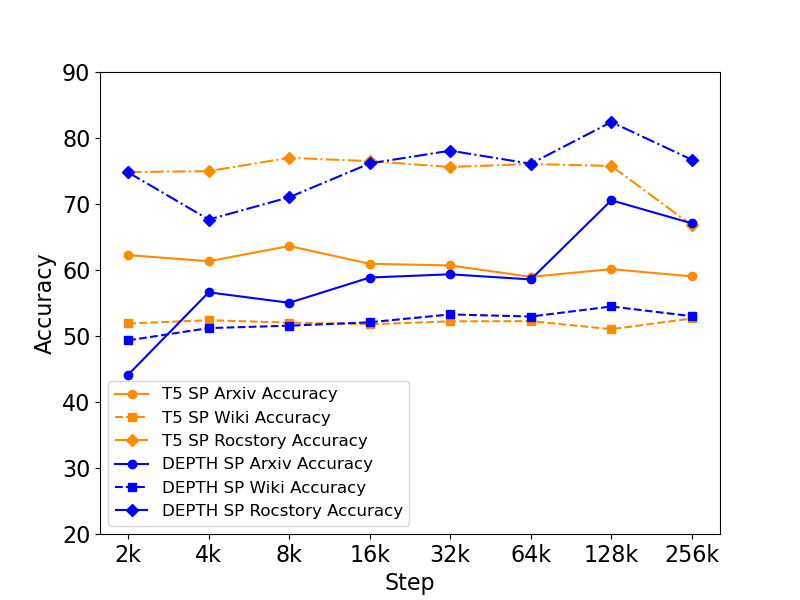}
        \caption{\footnotesize Sentence Permutation (SP) CPT}
        \label{fig:sp_cpt}
    \end{subfigure}
    \hfill
    \begin{subfigure}[b]{0.48\linewidth}
        \includegraphics[width=\linewidth]{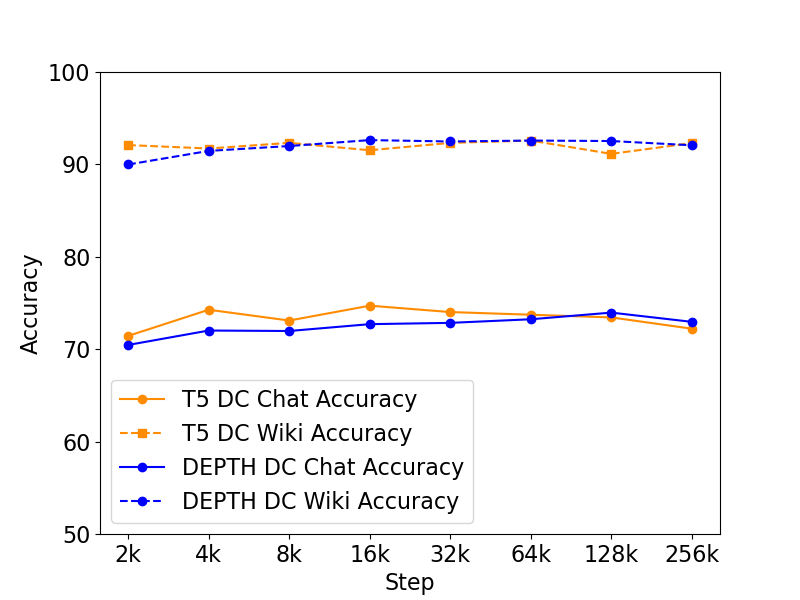}
        \caption{\footnotesize Discourse Coherence (DC) CPT}
        \label{fig:dc_cpt}
    \end{subfigure}
    \end{minipage}

    \caption{\small DiscoEval results for DEPTH and T5 models. Top row: From Scratch (FS), Bottom row: From Pretrained (CPT).}
    \label{fig:combined_discoeval_results}
\end{figure*}

We find that in the FS setting DEPTH consistently outperforms T5 across DC tasks, indicating its robustness in understanding discourse \footnote{In particular, sentence-level discourse relations, as discussed in \citet{jernite2017discoursebasedobjectivesfastunsupervised}} (Figure \ref{fig:combined_discoeval_results}, top row). This suggests DEPTH's pre-training objective is particularly beneficial for tasks that require a deep understanding of narrative structures (both in conversations as in DC-Chat, and more formal and informative texts as in DC-Wiki). We note that between steps 32k and 64k, DEPTH experienced a large positive boost in performance on DC-Wiki, perhaps indicative of an emergence \citep{wei2022emergent} of discourse understanding during this phase of pre-training. For T5, we found that the model struggled to learn the SP-Arxiv task, achieving random-guess accuracy late in pre-training. However, in SP-Wiki and SP-Rocstory, T5 improves in performances between steps 64k and 128k, perhaps indicating an emergent ability occurring within this timeframe. We report our full results on DiscoEval in Appendix \ref{sec:discoeval_results_appendix}.

\begin{table}[h!]
    \centering
    \vspace{1pt}
    \begin{tabular}{c c c}
    \toprule
        \textbf{Model} & 
        \textbf{SP} &
        \textbf{DC} \\
        
        \midrule
        
        RoBERTa-Base & 38.7 & 58.4 \\
        
        \midrule
        
        BERT-Base & 53.1 & 58.9 \\
        
        BERT-Large & 53.8 & 59.6 \\
        
        \midrule
        
        CONPONO & 60.7 & 72.9 \\
        
        \midrule
        
        SLM (1M)  & 72.4 & 75.4 \\
        SLM (3M) & \textbf{73.4} & 76.1 \\
        
        \midrule
        
        T5-Base & 58.1 & 80.5 \\
        
        T5-FS & 40.91 & 63.31 \\
        
        T5-CPT & 59.48 & \underline{82.27} \\
        
        \midrule
        
        DEPTH-FS & 55.45 & 76.22 \\
        DEPTH-CPT & \underline{65.59} & \textbf{82.49} \\
        
    \bottomrule
    \end{tabular}
    \caption{Comparison of various models on the SP and DC tasks within DiscoEval. All models aside from T5 and DEPTH and encoder-only models trained with discourse-oriented objectives. }
    \label{tab:discoeval_comparison}
\end{table}

While DEPTH outperformed other models in DC tasks, it failed to reach a high performance level on SP tasks (under-performing relative to SLM, as seen in Table \ref{tab:discoeval_comparison}).
This problem stems already from the pre-training stage, where DEPTH's sentence un-shuffling accuracy is relatively low ($\leq5\%$ accuracy on shuffled sentence tokens;
see Appendix \ref{sec:depth_loss_decomposition} for additional details).
This highlights the complexity of sentence un-shuffling relative to older discourse objectives like NSP and SOP. 
Surprisingly, while this task was challenging for DEPTH, SLM reported strong performance on sentence un-shuffling. SLM used a dedicated pointer-generator network that consists of a shallow DNN. This module ``points'' to one of at most $k$ sentences as it iterates over a target sequence consisting of \textbf{only} sentence tokens. Also, SLM's non-sentence tokens cannot observe sentence-level tokens as part of the reconstruction loss, avoiding a potential ``distraction'' in their task.

\subsection{NI fine-tuning}

\begin{figure*}[t]
    \centering
    \begin{minipage}{0.8\textwidth}
    \begin{subfigure}[b]{0.48\linewidth}
        \includegraphics[width=\linewidth]{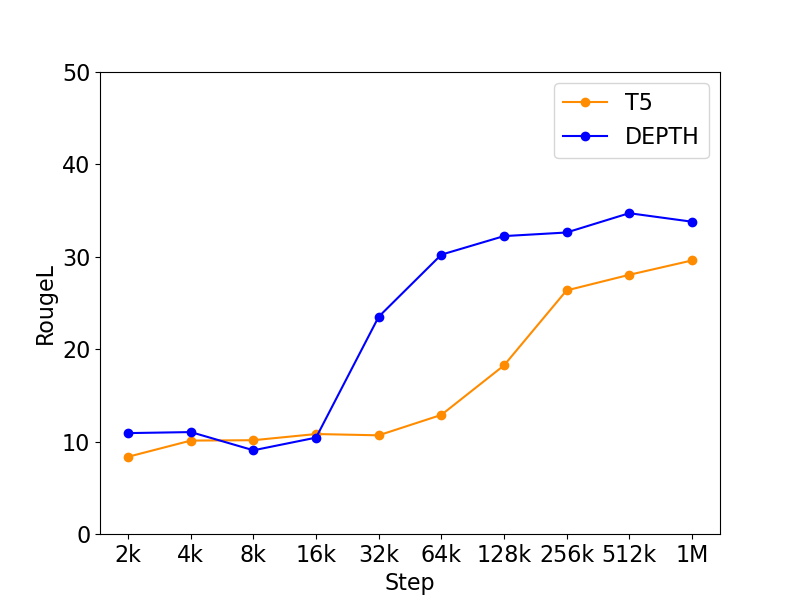}
        \caption{\footnotesize Natural Instructions (NI) FS}
        \label{fig:ni_fs}
    \end{subfigure}
    \hfill
    \begin{subfigure}[b]{0.48\linewidth}
        \includegraphics[width=\linewidth]{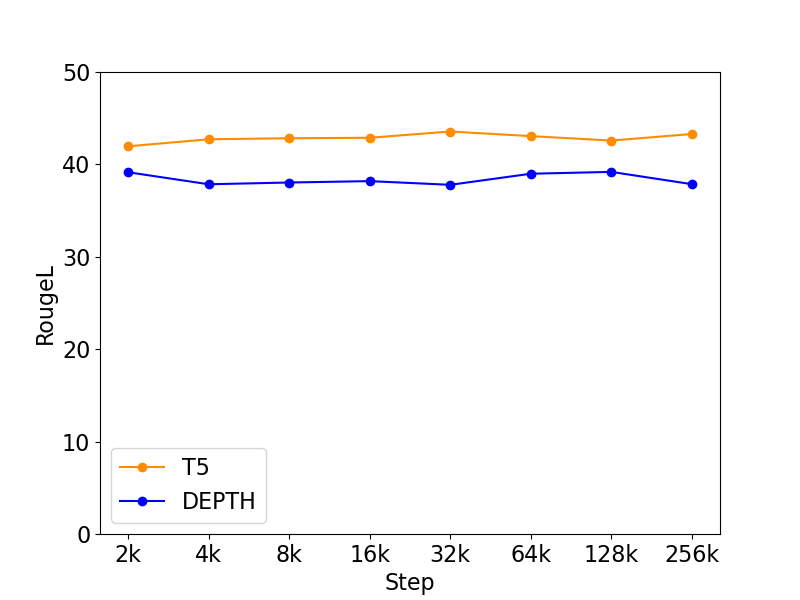}
        \caption{\footnotesize Natural Instructions (NI) CPT}
        \label{fig:ni_cpt}
    \end{subfigure}
    \end{minipage}
    \caption{\small NI results for DEPTH and T5 models.}
    \label{fig:combined_ni_results}
\end{figure*}

In the FS setting (Figure \ref{fig:ni_fs}), we observe that DEPTH outperforms T5 in the NI benchmark, with a notable leap in performance between steps 16k and 32k. T5, by comparison, only improves significantly after step 64k, and obtains worse performance than DEPTH by the end of training. 
However, in the CPT setting (Figure \ref{fig:ni_cpt}), DEPTH's pre-training appears to hinder downstream performance compared to T5, possibly due to the domain shift from T5's pre-training task to DEPTH's pre-training task, which involves learning from shuffled inputs. We present a more complete analysis of these results in Appendix \ref{sec:ni_results_appendix}.

\subsection{Error Analysis}

We performed error analysis on the DiscoEval benchmark to better understand the nature of discourse errors that DEPTH and T5 made. For the SP task, we show in Table \ref{tab:sp_error_analysis} that DEPTH made more reasonable mistakes than T5. For example, in SP-Arxiv FS, 23\% of DEPTH's mistakes were reasonable, relative to 7\% by T5). We define ``reasonable'' mistakes as incorrect predictions that would have still resulted in a coherent sentence ordering. Both models struggled with pronoun resolution, which frequently led to incorrect predictions (accounting for 10-30\% of all predictions we observed). T5-FS, in particular, often failed to recognize when a removed sentence should come first, a mistake largely resolved in T5-CPT. We note that some examples correctly predicted by FS models were incorrectly predicted by CPT models, and vice versa.

In the DC task, we noted a significant number of incorrectly formatted predictions (e.g., ``cooherent'' rather than ``coherent''), especially in the DC-Chat subset. Each of these incorrectly formatted predictions, when adjusted to a correctly formatted prediction, were incorrect (e.g., an example that the model predicted ``cooherent'' is labeled ``incoherent''). We show in Table \ref{tab:dc_prediction_types}, that DEPTH-FS was incorrect in DC-Chat examples that humans might find ambiguous (i.e., replacing a random sentence leaves the resulting passage coherent), reinforcing its strength in handling more complex discourse structures. We discuss this further in Appendix \ref{sec:error_analysis_appendix}.

\section{Related work}

The potential of encoder-decoder architectures in today's NLP landscape cannot be overstated. These architectures dominate context-heavy tasks ranging from translation \cite{cohere2024aya, xue-etal-2021-mt5, Tay2022UL2UL} to summarization \cite{zhang2020pegasus, guo-etal-2022-longt5, Tay2022UL2UL}, and even following instructions across diverse domains \cite{ext5_aribandi2022ext, wei2022finetuned, chung2024scaling}. Like their decoder-only counterparts, encoder-decoders are able to accommodate long inputs \citep{guo-etal-2022-longt5}, and scale effectively effectively as a function of model size and training data \citep{2024PileT5}. \citet{ormazabal2024reka} released a series of encoder-decoder models, where their dense 21B parameter model outperformed all models of its size in the lmsys benchmark \citep{zheng2023lmsys}.%
\footnote{This Reka model is competitve with mixtral 8x22b \citep{jiang2024mixtral} (which was trained with significantly more parameters using a mixture-of-experts architecture).} Encoder-deocder models are also strong multi-modal learners \citep{ormazabal2024reka, wu2023wav2seq, dosovitskiy2021an, scaling_vision_transformers}. When scaled sufficiently, encoder-decoders like Reca-Core may be competitive with state of the art models like GPT-4 \citep{openai2023gpt4}, Gemini \cite{geminiteam2023gemini}, and Claude-3.

While specialized encoder-decoder models such as PEGASUS \citep{zhang2020pegasus}, DialogVED \citep{chen-etal-2022-dialogved}, and a multi-party dialogue pre-training model \citep{li-etal-2023-pre-training-multi-party-dialogue} demonstrate the value of discourse-oriented tasks for encoder-decoder models, they have limited utility for broader tasks. Long-T5 \citep{guo-etal-2022-longt5} and UL2 \citep{Tay2022UL2UL} improved the ability of encoder-decoders to handle long contexts, but did not explicitly tackle discourse understanding. Flan-T5 \citep{wei2022finetuned} and Ex-T5 \citep{ext5_aribandi2022ext} demonstrated the applicability of encoder-decoders across a variety of tasks, including ones that are heavily discourse dependent. However, these models depend on a vast yet costly annotated dataset to learn human preferences.
Finally, BART \citep{lewis-etal-2020-bart} is an encoder-decoder which leverages sentence shuffling during pre-training, but does not train dedicated hierarchical representations for sentences (essentially behaving like a DEPTH model without sentence-tokens, and without attention-mask induced hierarchy).

\section{Limitations}
\label{sec:limitations}

Given our lack of computational resources (Appendix \ref{sec:computational_resources}), we were not able to pre-train our models with a batch size that would allow an aggressive learning rate like that used in \citet{raffel2020exploring_t5}'s T5 (see Appendix \ref{sec:replicating_t5} for additional details). We also pre-trained on substantially fewer tokens than T5. As a result, our model converges to a worse loss during pre-training, and performs worse on downstream tasks. We also lacked computational resources to compute confidence intervals or statistical significance for our downstream experiments.\footnote{Running a single downstream experiment on MNLI takes $5$-$7.5$ hours. We run $\approx120$ experiments for each of 10 benchmarks, and do not have the capacity to repeat experiments $\geq5\times$ to obtain statistical significance. }

Encoder-decoder LMs have fewer tools available for computationally efficient pre-training. For example, FlashAttention \citep{dao2022flashattention, dao2023flashattention2}, which provides a massive training speedup, is not available for encoder-decoder models. 
It is therefore difficult to create scalable pre-training experiments with new encoder-decoder architectures and objectives. 

\section{Conclusions and future work}

DEPTH's new pre-training objective and hierarchical representations complement efforts to scale model size, parallelize architectures, and acquire high quality data for pre-training. Despite training over fewer tokens, DEPTH significantly outperformed T5 both during pre-training and during fine-tuning.
DEPTH's remarkably efficient learning and downstream performance on discourse oriented tasks underscore the importance of discourse-oriented pre-training. 

Looking forward, the application of DEPTH to RAG \citep{Lewis2020RetrievalAugmentedGF}, especially over sentence ``chunks'', presents an exciting avenue for future research. Additionally, extending DEPTH's pre-training objectives to encompass higher-level discourse units---such as paragraphs, chapters, and whole documents---offers further flexibility emerging hierarchical RAG systems \citep{chen2024hiqa}. Moreover, conducting further experiments with larger DEPTH models is helpful for understanding the scalability of discourse-focused training objectives. Such investigations could reveal whether the promising capabilities observed in DEPTH are amplified with increased model capacity. Finally, sentence-level pre-training tasks such as next-sentence prediction (as in \citet{krishna-etal-2022-rankgen} and \citet{zhang2020pegasus}) may prove powerful alternatives to sentence un-shuffling. 


\section*{Acknowledgments}
This research was supported by the Israel Science
Foundation (grant No.\ 448/20), an Azrieli Foundation
Early Career Faculty Fellowship, and an AI Alignment grant from Open Philanthropy.

\bibliography{custom}

\begin{thebibliography}{83}
\providecommand{\natexlab}[1]{#1}

\bibitem[{Ankner et~al.(2024)Ankner, Saphra, Blalock, Frankle, and Leavitt}]{ankner-etal-2024-dynamic}
Zachary Ankner, Naomi Saphra, Davis Blalock, Jonathan Frankle, and Matthew Leavitt. 2024.
\newblock \href {https://aclanthology.org/2024.eacl-short.42/} {Dynamic masking rate schedules for {MLM} pretraining}.
\newblock In \emph{Proceedings of the 18th Conference of the European Chapter of the Association for Computational Linguistics (Volume 2: Short Papers)}, pages 477--487, St. Julian{'}s, Malta. Association for Computational Linguistics.

\bibitem[{Aribandi et~al.(2022)Aribandi, Tay, Schuster, Rao, Zheng, Mehta, Zhuang, Tran, Bahri, Ni, Gupta, Hui, Ruder, and Metzler}]{ext5_aribandi2022ext}
Vamsi Aribandi, Yi~Tay, Tal Schuster, Jinfeng Rao, Huaixiu~Steven Zheng, Sanket~Vaibhav Mehta, Honglei Zhuang, Vinh~Q. Tran, Dara Bahri, Jianmo Ni, Jai Gupta, Kai Hui, Sebastian Ruder, and Donald Metzler. 2022.
\newblock \href {https://openreview.net/forum?id=Vzh1BFUCiIX} {Ext5: Towards extreme multi-task scaling for transfer learning}.
\newblock In \emph{International Conference on Learning Representations}.

\bibitem[{Bai et~al.(2022)Bai, Kadavath, Kundu, Askell, Kernion, Jones, Chen, Goldie, Mirhoseini, McKinnon, Chen, Olsson, Olah, Hernandez, Drain, Ganguli, Li, Tran-Johnson, Perez, Kerr, Mueller, Ladish, Landau, Ndousse, Lukosuite, Lovitt, Sellitto, Elhage, Schiefer, Mercado, DasSarma, Lasenby, Larson, Ringer, Johnston, Kravec, Showk, Fort, Lanham, Telleen-Lawton, Conerly, Henighan, Hume, Bowman, Hatfield-Dodds, Mann, Amodei, Joseph, McCandlish, Brown, and Kaplan}]{bai2022constitutional}
Yuntao Bai, Saurav Kadavath, Sandipan Kundu, Amanda Askell, Jackson Kernion, Andy Jones, Anna Chen, Anna Goldie, Azalia Mirhoseini, Cameron McKinnon, Carol Chen, Catherine Olsson, Christopher Olah, Danny Hernandez, Dawn Drain, Deep Ganguli, Dustin Li, Eli Tran-Johnson, Ethan Perez, Jamie Kerr, Jared Mueller, Jeffrey Ladish, Joshua Landau, Kamal Ndousse, Kamile Lukosuite, Liane Lovitt, Michael Sellitto, Nelson Elhage, Nicholas Schiefer, Noemi Mercado, Nova DasSarma, Robert Lasenby, Robin Larson, Sam Ringer, Scott Johnston, Shauna Kravec, Sheer~El Showk, Stanislav Fort, Tamera Lanham, Timothy Telleen-Lawton, Tom Conerly, Tom Henighan, Tristan Hume, Samuel~R. Bowman, Zac Hatfield-Dodds, Ben Mann, Dario Amodei, Nicholas Joseph, Sam McCandlish, Tom Brown, and Jared Kaplan. 2022.
\newblock \href {https://arxiv.org/abs/2212.08073} {Constitutional ai: Harmlessness from ai feedback}.
\newblock \emph{Preprint}, arXiv:2212.08073.

\bibitem[{Biderman et~al.(2023)Biderman, Prashanth, Sutawika, Schoelkopf, Anthony, Purohit, and Raff}]{pythia_biderman2023emergent}
Stella Biderman, USVSN~Sai Prashanth, Lintang Sutawika, Hailey Schoelkopf, Quentin Anthony, Shivanshu Purohit, and Edward Raff. 2023.
\newblock Emergent and predictable memorization in large language models.

\bibitem[{Bird and Loper(2004)}]{bird-loper-2004-nltk}
Steven Bird and Edward Loper. 2004.
\newblock \href {https://aclanthology.org/P04-3031} {{NLTK}: The natural language toolkit}.
\newblock In \emph{Proceedings of the {ACL} Interactive Poster and Demonstration Sessions}, pages 214--217, Barcelona, Spain. Association for Computational Linguistics.

\bibitem[{Brown et~al.(2020)Brown, Mann, Ryder, Subbiah, Kaplan, Dhariwal, Neelakantan, Shyam, Sastry, Askell, Agarwal, Herbert-Voss, Krueger, Henighan, Child, Ramesh, Ziegler, Wu, Winter, Hesse, Chen, Sigler, Litwin, Gray, Chess, Clark, Berner, McCandlish, Radford, Sutskever, and Amodei}]{gpt3}
Tom Brown, Benjamin Mann, Nick Ryder, Melanie Subbiah, Jared~D Kaplan, Prafulla Dhariwal, Arvind Neelakantan, Pranav Shyam, Girish Sastry, Amanda Askell, Sandhini Agarwal, Ariel Herbert-Voss, Gretchen Krueger, Tom Henighan, Rewon Child, Aditya Ramesh, Daniel Ziegler, Jeffrey Wu, Clemens Winter, Chris Hesse, Mark Chen, Eric Sigler, Mateusz Litwin, Scott Gray, Benjamin Chess, Jack Clark, Christopher Berner, Sam McCandlish, Alec Radford, Ilya Sutskever, and Dario Amodei. 2020.
\newblock \href {https://proceedings.neurips.cc/paper_files/paper/2020/file/1457c0d6bfcb4967418bfb8ac142f64a-Paper.pdf} {Language models are few-shot learners}.
\newblock In \emph{Advances in Neural Information Processing Systems}, volume~33, pages 1877--1901. Curran Associates, Inc.

\bibitem[{Chakrabarty et~al.(2019)Chakrabarty, Hidey, Muresan, McKeown, and Hwang}]{chakrabarty-etal-2019-ampersand}
Tuhin Chakrabarty, Christopher Hidey, Smaranda Muresan, Kathy McKeown, and Alyssa Hwang. 2019.
\newblock \href {https://doi.org/10.18653/v1/D19-1291} {{AMPERSAND}: Argument mining for {PERS}u{A}sive o{N}line discussions}.
\newblock In \emph{Proceedings of the 2019 Conference on Empirical Methods in Natural Language Processing and the 9th International Joint Conference on Natural Language Processing (EMNLP-IJCNLP)}, pages 2933--2943, Hong Kong, China. Association for Computational Linguistics.

\bibitem[{Chen et~al.(2019)Chen, Chu, and Gimpel}]{chen-etal-2019-evaluation-discoeval}
Mingda Chen, Zewei Chu, and Kevin Gimpel. 2019.
\newblock \href {https://doi.org/10.18653/v1/D19-1060} {Evaluation benchmarks and learning criteria for discourse-aware sentence representations}.
\newblock In \emph{Proceedings of the 2019 Conference on Empirical Methods in Natural Language Processing and the 9th International Joint Conference on Natural Language Processing (EMNLP-IJCNLP)}, pages 649--662, Hong Kong, China. Association for Computational Linguistics.

\bibitem[{Chen et~al.(2022)Chen, Gong, Wang, Yao, Qi, Wei, Hu, Zhou, Mao, Chen, Cheng, and Duan}]{chen-etal-2022-dialogved}
Wei Chen, Yeyun Gong, Song Wang, Bolun Yao, Weizhen Qi, Zhongyu Wei, Xiaowu Hu, Bartuer Zhou, Yi~Mao, Weizhu Chen, Biao Cheng, and Nan Duan. 2022.
\newblock \href {https://doi.org/10.18653/v1/2022.acl-long.333} {{D}ialog{VED}: A pre-trained latent variable encoder-decoder model for dialog response generation}.
\newblock In \emph{Proceedings of the 60th Annual Meeting of the Association for Computational Linguistics (Volume 1: Long Papers)}, pages 4852--4864, Dublin, Ireland. Association for Computational Linguistics.

\bibitem[{Chen et~al.(2024)Chen, Gao, Song, and Tan}]{chen2024hiqa}
Xinyue Chen, Pengyu Gao, Jiangjiang Song, and Xiaoyang Tan. 2024.
\newblock \href {https://arxiv.org/abs/2402.01767} {Hiqa: A hierarchical contextual augmentation rag for massive documents qa}.
\newblock \emph{Preprint}, arXiv:2402.01767.

\bibitem[{Chung et~al.(2024)Chung, Hou, Longpre, Zoph, Tay, Fedus, Li, Wang, Dehghani, Brahma et~al.}]{chung2024scaling}
Hyung~Won Chung, Le~Hou, Shayne Longpre, Barret Zoph, Yi~Tay, William Fedus, Yunxuan Li, Xuezhi Wang, Mostafa Dehghani, Siddhartha Brahma, et~al. 2024.
\newblock Scaling instruction-finetuned language models.
\newblock \emph{Journal of Machine Learning Research}, 25(70):1--53.

\bibitem[{Computer(2023)}]{together2023redpajama}
Together Computer. 2023.
\newblock \href {https://github.com/togethercomputer/RedPajama-Data} {Redpajama: an open dataset for training large language models}.

\bibitem[{Dao(2023)}]{dao2023flashattention2}
Tri Dao. 2023.
\newblock Flash{A}ttention-2: Faster attention with better parallelism and work partitioning.

\bibitem[{Dao et~al.(2022)Dao, Fu, Ermon, Rudra, and R{\'e}}]{dao2022flashattention}
Tri Dao, Daniel~Y. Fu, Stefano Ermon, Atri Rudra, and Christopher R{\'e}. 2022.
\newblock Flash{A}ttention: Fast and memory-efficient exact attention with {IO}-awareness.
\newblock In \emph{Advances in Neural Information Processing Systems}.

\bibitem[{Devlin et~al.(2019)Devlin, Chang, Lee, and Toutanova}]{devlin-etal-2019-bert}
Jacob Devlin, Ming-Wei Chang, Kenton Lee, and Kristina Toutanova. 2019.
\newblock \href {https://doi.org/10.18653/v1/N19-1423} {{BERT}: Pre-training of deep bidirectional transformers for language understanding}.
\newblock In \emph{Proceedings of the 2019 Conference of the North {A}merican Chapter of the Association for Computational Linguistics: Human Language Technologies, Volume 1 (Long and Short Papers)}, pages 4171--4186, Minneapolis, Minnesota. Association for Computational Linguistics.

\bibitem[{Ding et~al.(2024)Ding, Wang, Paolini, Kumar, Deoras, Roth, and Soatto}]{ding2024fewertruncations}
Hantian Ding, Zijian Wang, Giovanni Paolini, Varun Kumar, Anoop Deoras, Dan Roth, and Stefano Soatto. 2024.
\newblock Fewer truncations improve language modeling.
\newblock \emph{arXiv preprint arXiv:2404.10830}.

\bibitem[{Dodge et~al.(2021)Dodge, Sap, Marasovi{\'c}, Agnew, Ilharco, Groeneveld, Mitchell, and Gardner}]{dodge-etal-2021-documenting_c4}
Jesse Dodge, Maarten Sap, Ana Marasovi{\'c}, William Agnew, Gabriel Ilharco, Dirk Groeneveld, Margaret Mitchell, and Matt Gardner. 2021.
\newblock \href {https://doi.org/10.18653/v1/2021.emnlp-main.98} {Documenting large webtext corpora: A case study on the colossal clean crawled corpus}.
\newblock In \emph{Proceedings of the 2021 Conference on Empirical Methods in Natural Language Processing}, pages 1286--1305, Online and Punta Cana, Dominican Republic. Association for Computational Linguistics.

\bibitem[{Dosovitskiy et~al.(2021)Dosovitskiy, Beyer, Kolesnikov, Weissenborn, Zhai, Unterthiner, Dehghani, Minderer, Heigold, Gelly, Uszkoreit, and Houlsby}]{dosovitskiy2021an}
Alexey Dosovitskiy, Lucas Beyer, Alexander Kolesnikov, Dirk Weissenborn, Xiaohua Zhai, Thomas Unterthiner, Mostafa Dehghani, Matthias Minderer, Georg Heigold, Sylvain Gelly, Jakob Uszkoreit, and Neil Houlsby. 2021.
\newblock \href {https://openreview.net/forum?id=YicbFdNTTy} {An image is worth 16x16 words: Transformers for image recognition at scale}.
\newblock In \emph{International Conference on Learning Representations}.

\bibitem[{Durmus et~al.(2019)Durmus, Ladhak, and Cardie}]{Durmus2019-pragmatic-discourse}
Esin Durmus, Faisal Ladhak, and Claire Cardie. 2019.
\newblock \href {https://api.semanticscholar.org/CorpusID:202768765} {The role of pragmatic and discourse context in determining argument impact}.
\newblock In \emph{Conference on Empirical Methods in Natural Language Processing}.

\bibitem[{Grattafiori et~al.(2024)Grattafiori, Dubey, Jauhri, Pandey, Kadian, Al-Dahle, Letman, Mathur, Schelten, Vaughan et~al.}]{grattafiori2024llama}
Aaron Grattafiori, Abhimanyu Dubey, Abhinav Jauhri, Abhinav Pandey, Abhishek Kadian, Ahmad Al-Dahle, Aiesha Letman, Akhil Mathur, Alan Schelten, Alex Vaughan, et~al. 2024.
\newblock The llama 3 herd of models.
\newblock \emph{arXiv preprint arXiv:2407.21783}.

\bibitem[{Groeneveld et~al.(2024)Groeneveld, Beltagy, Walsh, Bhagia, Kinney, Tafjord, Jha, Ivison, Magnusson, Wang, Arora, Atkinson, Authur, Chandu, Cohan, Dumas, Elazar, Gu, Hessel, Khot, Merrill, Morrison, Muennighoff, Naik, Nam, Peters, Pyatkin, Ravichander, Schwenk, Shah, Smith, Subramani, Wortsman, Dasigi, Lambert, Richardson, Dodge, Lo, Soldaini, Smith, and Hajishirzi}]{Groeneveld2023OLMo}
Dirk Groeneveld, Iz~Beltagy, Pete Walsh, Akshita Bhagia, Rodney Kinney, Oyvind Tafjord, Ananya~Harsh Jha, Hamish Ivison, Ian Magnusson, Yizhong Wang, Shane Arora, David Atkinson, Russell Authur, Khyathi Chandu, Arman Cohan, Jennifer Dumas, Yanai Elazar, Yuling Gu, Jack Hessel, Tushar Khot, William Merrill, Jacob Morrison, Niklas Muennighoff, Aakanksha Naik, Crystal Nam, Matthew~E. Peters, Valentina Pyatkin, Abhilasha Ravichander, Dustin Schwenk, Saurabh Shah, Will Smith, Nishant Subramani, Mitchell Wortsman, Pradeep Dasigi, Nathan Lambert, Kyle Richardson, Jesse Dodge, Kyle Lo, Luca Soldaini, Noah~A. Smith, and Hannaneh Hajishirzi. 2024.
\newblock Olmo: Accelerating the science of language models.
\newblock \emph{Preprint}.

\bibitem[{Guo et~al.(2022)Guo, Ainslie, Uthus, Ontanon, Ni, Sung, and Yang}]{guo-etal-2022-longt5}
Mandy Guo, Joshua Ainslie, David Uthus, Santiago Ontanon, Jianmo Ni, Yun-Hsuan Sung, and Yinfei Yang. 2022.
\newblock \href {https://doi.org/10.18653/v1/2022.findings-naacl.55} {{L}ong{T}5: {E}fficient text-to-text transformer for long sequences}.
\newblock In \emph{Findings of the Association for Computational Linguistics: NAACL 2022}, pages 724--736, Seattle, United States. Association for Computational Linguistics.

\bibitem[{Hidey et~al.(2017)Hidey, Musi, Hwang, Muresan, and McKeown}]{hidey-etal-2017-analyzing}
Christopher Hidey, Elena Musi, Alyssa Hwang, Smaranda Muresan, and Kathy McKeown. 2017.
\newblock \href {https://doi.org/10.18653/v1/W17-5102} {Analyzing the semantic types of claims and premises in an online persuasive forum}.
\newblock In \emph{Proceedings of the 4th Workshop on Argument Mining}, pages 11--21, Copenhagen, Denmark. Association for Computational Linguistics.

\bibitem[{Hua et~al.(2023)Hua, Deng, and McKeown}]{hua-etal-2023-improving-dialogue-summarization}
Yilun Hua, Zhaoyuan Deng, and Kathleen McKeown. 2023.
\newblock \href {https://doi.org/10.18653/v1/2023.findings-acl.871} {Improving long dialogue summarization with semantic graph representation}.
\newblock In \emph{Findings of the Association for Computational Linguistics: ACL 2023}, pages 13851--13883, Toronto, Canada. Association for Computational Linguistics.

\bibitem[{Iter et~al.(2020)Iter, Guu, Lansing, and Jurafsky}]{iter-etal-2020-pretraining-compono}
Dan Iter, Kelvin Guu, Larry Lansing, and Dan Jurafsky. 2020.
\newblock \href {https://doi.org/10.18653/v1/2020.acl-main.439} {Pretraining with contrastive sentence objectives improves discourse performance of language models}.
\newblock In \emph{Proceedings of the 58th Annual Meeting of the Association for Computational Linguistics}, pages 4859--4870, Online. Association for Computational Linguistics.

\bibitem[{Jernite et~al.(2017)Jernite, Bowman, and Sontag}]{jernite2017discoursebasedobjectivesfastunsupervised}
Yacine Jernite, Samuel~R. Bowman, and David Sontag. 2017.
\newblock \href {https://arxiv.org/abs/1705.00557} {Discourse-based objectives for fast unsupervised sentence representation learning}.
\newblock \emph{Preprint}, arXiv:1705.00557.

\bibitem[{Jiang et~al.(2023)Jiang, Sablayrolles, Mensch, Bamford, Chaplot, de~las Casas, Bressand, Lengyel, Lample, Saulnier, Lavaud, Lachaux, Stock, Scao, Lavril, Wang, Lacroix, and Sayed}]{jiang2023mistral7b}
Albert~Q. Jiang, Alexandre Sablayrolles, Arthur Mensch, Chris Bamford, Devendra~Singh Chaplot, Diego de~las Casas, Florian Bressand, Gianna Lengyel, Guillaume Lample, Lucile Saulnier, Lélio~Renard Lavaud, Marie-Anne Lachaux, Pierre Stock, Teven~Le Scao, Thibaut Lavril, Thomas Wang, Timothée Lacroix, and William~El Sayed. 2023.
\newblock \href {https://arxiv.org/abs/2310.06825} {Mistral 7b}.
\newblock \emph{Preprint}, arXiv:2310.06825.

\bibitem[{Jiang et~al.(2024)Jiang, Sablayrolles, Roux, Mensch, Savary, Bamford, Chaplot, Casas, Hanna, Bressand et~al.}]{jiang2024mixtral}
Albert~Q Jiang, Alexandre Sablayrolles, Antoine Roux, Arthur Mensch, Blanche Savary, Chris Bamford, Devendra~Singh Chaplot, Diego de~las Casas, Emma~Bou Hanna, Florian Bressand, et~al. 2024.
\newblock Mixtral of experts.
\newblock \emph{arXiv preprint arXiv:2401.04088}.

\bibitem[{Joshi et~al.(2020)Joshi, Chen, Liu, Weld, Zettlemoyer, and Levy}]{joshi-etal-2020-spanbert}
Mandar Joshi, Danqi Chen, Yinhan Liu, Daniel~S. Weld, Luke Zettlemoyer, and Omer Levy. 2020.
\newblock \href {https://doi.org/10.1162/tacl_a_00300} {{S}pan{BERT}: Improving pre-training by representing and predicting spans}.
\newblock \emph{Transactions of the Association for Computational Linguistics}, 8:64--77.

\bibitem[{Katz et~al.(2024)Katz, Ringel, Romano, and Wolf}]{katz2024segmentbasedattentionmaskinggpts}
Shahar Katz, Liran Ringel, Yaniv Romano, and Lior Wolf. 2024.
\newblock \href {https://arxiv.org/abs/2412.18487} {Segment-based attention masking for gpts}.
\newblock \emph{Preprint}, arXiv:2412.18487.

\bibitem[{Krell et~al.(2021)Krell, Kosec, Perez, and Fitzgibbon}]{krell2021efficientsequencepacking}
Mario~Michael Krell, Matej Kosec, Sergio~P Perez, and Andrew Fitzgibbon. 2021.
\newblock Efficient sequence packing without cross-contamination: Accelerating large language models without impacting performance.
\newblock \emph{arXiv preprint arXiv:2107.02027}.

\bibitem[{Krishna et~al.(2022)Krishna, Chang, Wieting, and Iyyer}]{krishna-etal-2022-rankgen}
Kalpesh Krishna, Yapei Chang, John Wieting, and Mohit Iyyer. 2022.
\newblock \href {https://doi.org/10.18653/v1/2022.emnlp-main.15} {{R}ank{G}en: Improving text generation with large ranking models}.
\newblock In \emph{Proceedings of the 2022 Conference on Empirical Methods in Natural Language Processing}, pages 199--232, Abu Dhabi, United Arab Emirates. Association for Computational Linguistics.

\bibitem[{Kudo and Richardson(2018)}]{kudo-richardson-2018-sentencepiece}
Taku Kudo and John Richardson. 2018.
\newblock \href {https://doi.org/10.18653/v1/D18-2012} {{S}entence{P}iece: A simple and language independent subword tokenizer and detokenizer for neural text processing}.
\newblock In \emph{Proceedings of the 2018 Conference on Empirical Methods in Natural Language Processing: System Demonstrations}, pages 66--71, Brussels, Belgium. Association for Computational Linguistics.

\bibitem[{Lan et~al.(2020)Lan, Chen, Goodman, Gimpel, Sharma, and Soricut}]{Lan2020ALBERT:}
Zhenzhong Lan, Mingda Chen, Sebastian Goodman, Kevin Gimpel, Piyush Sharma, and Radu Soricut. 2020.
\newblock \href {https://openreview.net/forum?id=H1eA7AEtvS} {Albert: A lite bert for self-supervised learning of language representations}.
\newblock In \emph{International Conference on Learning Representations}.

\bibitem[{Lee et~al.(2020)Lee, Hudson, Lee, and Manning}]{lee-etal-2020-slm}
Haejun Lee, Drew~A. Hudson, Kangwook Lee, and Christopher~D. Manning. 2020.
\newblock \href {https://doi.org/10.18653/v1/2020.emnlp-main.120} {{SLM}: Learning a discourse language representation with sentence unshuffling}.
\newblock In \emph{Proceedings of the 2020 Conference on Empirical Methods in Natural Language Processing (EMNLP)}, pages 1551--1562, Online. Association for Computational Linguistics.

\bibitem[{Levine et~al.(2020)Levine, Lenz, Lieber, Abend, Leyton-Brown, Tennenholtz, and Shoham}]{levine2020pmi}
Yoav Levine, Barak Lenz, Opher Lieber, Omri Abend, Kevin Leyton-Brown, Moshe Tennenholtz, and Yoav Shoham. 2020.
\newblock Pmi-masking: Principled masking of correlated spans.
\newblock In \emph{International Conference on Learning Representations}.

\bibitem[{Lewis et~al.(2020{\natexlab{a}})Lewis, Liu, Goyal, Ghazvininejad, Mohamed, Levy, Stoyanov, and Zettlemoyer}]{lewis-etal-2020-bart}
Mike Lewis, Yinhan Liu, Naman Goyal, Marjan Ghazvininejad, Abdelrahman Mohamed, Omer Levy, Veselin Stoyanov, and Luke Zettlemoyer. 2020{\natexlab{a}}.
\newblock \href {https://doi.org/10.18653/v1/2020.acl-main.703} {{BART}: Denoising sequence-to-sequence pre-training for natural language generation, translation, and comprehension}.
\newblock In \emph{Proceedings of the 58th Annual Meeting of the Association for Computational Linguistics}, pages 7871--7880, Online. Association for Computational Linguistics.

\bibitem[{Lewis et~al.(2020{\natexlab{b}})Lewis, Perez, Piktus, Petroni, Karpukhin, Goyal, Kuttler, Lewis, tau Yih, Rockt{\"a}schel, Riedel, and Kiela}]{Lewis2020RetrievalAugmentedGF}
Patrick Lewis, Ethan Perez, Aleksandara Piktus, Fabio Petroni, Vladimir Karpukhin, Naman Goyal, Heinrich Kuttler, Mike Lewis, Wen tau Yih, Tim Rockt{\"a}schel, Sebastian Riedel, and Douwe Kiela. 2020{\natexlab{b}}.
\newblock \href {https://api.semanticscholar.org/CorpusID:218869575} {Retrieval-augmented generation for knowledge-intensive nlp tasks}.
\newblock \emph{ArXiv}, abs/2005.11401.

\bibitem[{Li et~al.(2023)Li, Huang, Bi, and Zhao}]{li-etal-2023-pre-training-multi-party-dialogue}
Yiyang Li, Xinting Huang, Wei Bi, and Hai Zhao. 2023.
\newblock \href {https://doi.org/10.18653/v1/2023.acl-long.533} {Pre-training multi-party dialogue models with latent discourse inference}.
\newblock In \emph{Proceedings of the 61st Annual Meeting of the Association for Computational Linguistics (Volume 1: Long Papers)}, pages 9584--9599, Toronto, Canada. Association for Computational Linguistics.

\bibitem[{Liu et~al.(2019)Liu, Ott, Goyal, Du, Joshi, Chen, Levy, Lewis, Zettlemoyer, and Stoyanov}]{liu2019roberta}
Yinhan Liu, Myle Ott, Naman Goyal, Jingfei Du, Mandar Joshi, Danqi Chen, Omer Levy, Mike Lewis, Luke Zettlemoyer, and Veselin Stoyanov. 2019.
\newblock \href {https://arxiv.org/abs/1907.11692} {Roberta: A robustly optimized bert pretraining approach}.
\newblock \emph{Preprint}, arXiv:1907.11692.

\bibitem[{Loshchilov and Hutter(2019)}]{loshchilov2018decoupledAdamW}
Ilya Loshchilov and Frank Hutter. 2019.
\newblock \href {https://openreview.net/forum?id=Bkg6RiCqY7} {Decoupled weight decay regularization}.
\newblock In \emph{International Conference on Learning Representations}.

\bibitem[{Maimon and Tsarfaty(2023{\natexlab{a}})}]{maimon-tsarfaty-2023-cohesentia}
Aviya Maimon and Reut Tsarfaty. 2023{\natexlab{a}}.
\newblock \href {https://doi.org/10.18653/v1/2023.emnlp-main.324} {{COHESENTIA}: A novel benchmark of incremental versus holistic assessment of coherence in generated texts}.
\newblock In \emph{Proceedings of the 2023 Conference on Empirical Methods in Natural Language Processing}, pages 5328--5343, Singapore. Association for Computational Linguistics.

\bibitem[{Maimon and Tsarfaty(2023{\natexlab{b}})}]{maimon-2023-coherence-assessment}
Aviya Maimon and Reut Tsarfaty. 2023{\natexlab{b}}.
\newblock \href {https://arxiv.org/abs/2310.00598} {A novel computational and modeling foundation for automatic coherence assessment}.
\newblock \emph{Preprint}, arXiv:2310.00598.

\bibitem[{Miltsakaki et~al.(2004)Miltsakaki, Prasad, Joshi, and Webber}]{miltsakaki-etal-2004-penn}
Eleni Miltsakaki, Rashmi Prasad, Aravind Joshi, and Bonnie Webber. 2004.
\newblock \href {http://www.lrec-conf.org/proceedings/lrec2004/pdf/618.pdf} {The {P}enn {D}iscourse {T}reebank}.
\newblock In \emph{Proceedings of the Fourth International Conference on Language Resources and Evaluation ({LREC}{'}04)}, Lisbon, Portugal. European Language Resources Association (ELRA).

\bibitem[{Mim et~al.(2021)Mim, Inoue, Reisert, Ouchi, and Inui}]{corruption-is-not-all-bad-2021}
Farjana~Sultana Mim, Naoya Inoue, Paul Reisert, Hiroki Ouchi, and Kentaro Inui. 2021.
\newblock \href {https://doi.org/10.1109/TASLP.2021.3088223} {Corruption is not all bad: Incorporating discourse structure into pre-training via corruption for essay scoring}.
\newblock \emph{IEEE/ACM Transactions on Audio, Speech, and Language Processing}, 29:2202--2215.

\bibitem[{Mishra et~al.(2022)Mishra, Khashabi, Baral, and Hajishirzi}]{mishra-etal-2022-cross}
Swaroop Mishra, Daniel Khashabi, Chitta Baral, and Hannaneh Hajishirzi. 2022.
\newblock \href {https://doi.org/10.18653/v1/2022.acl-long.244} {Cross-task generalization via natural language crowdsourcing instructions}.
\newblock In \emph{Proceedings of the 60th Annual Meeting of the Association for Computational Linguistics (Volume 1: Long Papers)}, pages 3470--3487, Dublin, Ireland. Association for Computational Linguistics.

\bibitem[{Muennighoff et~al.(2023)Muennighoff, Rush, Barak, Scao, Tazi, Piktus, Pyysalo, Wolf, and Raffel}]{muennighoff2023scaling}
Niklas Muennighoff, Alexander~M Rush, Boaz Barak, Teven~Le Scao, Nouamane Tazi, Aleksandra Piktus, Sampo Pyysalo, Thomas Wolf, and Colin Raffel. 2023.
\newblock \href {https://openreview.net/forum?id=j5BuTrEj35} {Scaling data-constrained language models}.
\newblock In \emph{Thirty-seventh Conference on Neural Information Processing Systems}.

\bibitem[{Nawrot(2023)}]{nawrot-2023-nanot5}
Piotr Nawrot. 2023.
\newblock \href {https://doi.org/10.18653/v1/2023.nlposs-1.11} {nano{T}5: Fast {\&} simple pre-training and fine-tuning of t5 models with limited resources}.
\newblock In \emph{Proceedings of the 3rd Workshop for Natural Language Processing Open Source Software (NLP-OSS 2023)}, pages 95--101, Singapore. Association for Computational Linguistics.

\bibitem[{OpenAI et~al.(2023)OpenAI, :, Achiam, Adler, Agarwal, Ahmad, Akkaya, Aleman, Almeida, Altenschmidt, Altman, Anadkat, Avila, Babuschkin, Balaji, Balcom, Baltescu, Bao, Bavarian, Belgum, Bello, Berdine, Bernadett-Shapiro, Berner, Bogdonoff, Boiko, Boyd, Brakman, Brockman, Brooks, Brundage, Button, Cai, Campbell, Cann, Carey, Carlson, Carmichael, Chan, Chang, Chantzis, Chen, Chen, Chen, Chen, Chen, Chess, Cho, Chu, Chung, Cummings, Currier, Dai, Decareaux, Degry, Deutsch, Deville, Dhar, Dohan, Dowling, Dunning, Ecoffet, Eleti, Eloundou, Farhi, Fedus, Felix, Fishman, Forte, Fulford, Gao, Georges, Gibson, Goel, Gogineni, Goh, Gontijo-Lopes, Gordon, Grafstein, Gray, Greene, Gross, Gu, Guo, Hallacy, Han, Harris, He, Heaton, Heidecke, Hesse, Hickey, Hickey, Hoeschele, Houghton, Hsu, Hu, Hu, Huizinga, Jain, Jain, Jang, Jiang, Jiang, Jin, Jin, Jomoto, Jonn, Jun, Kaftan, Łukasz Kaiser, Kamali, Kanitscheider, Keskar, Khan, Kilpatrick, Kim, Kim, Kim, Kirchner, Kiros, Knight, Kokotajlo, Łukasz Kondraciuk,
  Kondrich, Konstantinidis, Kosic, Krueger, Kuo, Lampe, Lan, Lee, Leike, Leung, Levy, Li, Lim, Lin, Lin, Litwin, Lopez, Lowe, Lue, Makanju, Malfacini, Manning, Markov, Markovski, Martin, Mayer, Mayne, McGrew, McKinney, McLeavey, McMillan, McNeil, Medina, Mehta, Menick, Metz, Mishchenko, Mishkin, Monaco, Morikawa, Mossing, Mu, Murati, Murk, Mély, Nair, Nakano, Nayak, Neelakantan, Ngo, Noh, Ouyang, O'Keefe, Pachocki, Paino, Palermo, Pantuliano, Parascandolo, Parish, Parparita, Passos, Pavlov, Peng, Perelman, de~Avila Belbute~Peres, Petrov, de~Oliveira~Pinto, Michael, Pokorny, Pokrass, Pong, Powell, Power, Power, Proehl, Puri, Radford, Rae, Ramesh, Raymond, Real, Rimbach, Ross, Rotsted, Roussez, Ryder, Saltarelli, Sanders, Santurkar, Sastry, Schmidt, Schnurr, Schulman, Selsam, Sheppard, Sherbakov, Shieh, Shoker, Shyam, Sidor, Sigler, Simens, Sitkin, Slama, Sohl, Sokolowsky, Song, Staudacher, Such, Summers, Sutskever, Tang, Tezak, Thompson, Tillet, Tootoonchian, Tseng, Tuggle, Turley, Tworek, Uribe, Vallone,
  Vijayvergiya, Voss, Wainwright, Wang, Wang, Wang, Ward, Wei, Weinmann, Welihinda, Welinder, Weng, Weng, Wiethoff, Willner, Winter, Wolrich, Wong, Workman, Wu, Wu, Wu, Xiao, Xu, Yoo, Yu, Yuan, Zaremba, Zellers, Zhang, Zhang, Zhao, Zheng, Zhuang, Zhuk, and Zoph}]{openai2023gpt4}
OpenAI, :, Josh Achiam, Steven Adler, Sandhini Agarwal, Lama Ahmad, Ilge Akkaya, Florencia~Leoni Aleman, Diogo Almeida, Janko Altenschmidt, Sam Altman, Shyamal Anadkat, Red Avila, Igor Babuschkin, Suchir Balaji, Valerie Balcom, Paul Baltescu, Haiming Bao, Mo~Bavarian, Jeff Belgum, Irwan Bello, Jake Berdine, Gabriel Bernadett-Shapiro, Christopher Berner, Lenny Bogdonoff, Oleg Boiko, Madelaine Boyd, Anna-Luisa Brakman, Greg Brockman, Tim Brooks, Miles Brundage, Kevin Button, Trevor Cai, Rosie Campbell, Andrew Cann, Brittany Carey, Chelsea Carlson, Rory Carmichael, Brooke Chan, Che Chang, Fotis Chantzis, Derek Chen, Sully Chen, Ruby Chen, Jason Chen, Mark Chen, Ben Chess, Chester Cho, Casey Chu, Hyung~Won Chung, Dave Cummings, Jeremiah Currier, Yunxing Dai, Cory Decareaux, Thomas Degry, Noah Deutsch, Damien Deville, Arka Dhar, David Dohan, Steve Dowling, Sheila Dunning, Adrien Ecoffet, Atty Eleti, Tyna Eloundou, David Farhi, Liam Fedus, Niko Felix, Simón~Posada Fishman, Juston Forte, Isabella Fulford, Leo Gao,
  Elie Georges, Christian Gibson, Vik Goel, Tarun Gogineni, Gabriel Goh, Rapha Gontijo-Lopes, Jonathan Gordon, Morgan Grafstein, Scott Gray, Ryan Greene, Joshua Gross, Shixiang~Shane Gu, Yufei Guo, Chris Hallacy, Jesse Han, Jeff Harris, Yuchen He, Mike Heaton, Johannes Heidecke, Chris Hesse, Alan Hickey, Wade Hickey, Peter Hoeschele, Brandon Houghton, Kenny Hsu, Shengli Hu, Xin Hu, Joost Huizinga, Shantanu Jain, Shawn Jain, Joanne Jang, Angela Jiang, Roger Jiang, Haozhun Jin, Denny Jin, Shino Jomoto, Billie Jonn, Heewoo Jun, Tomer Kaftan, Łukasz Kaiser, Ali Kamali, Ingmar Kanitscheider, Nitish~Shirish Keskar, Tabarak Khan, Logan Kilpatrick, Jong~Wook Kim, Christina Kim, Yongjik Kim, Hendrik Kirchner, Jamie Kiros, Matt Knight, Daniel Kokotajlo, Łukasz Kondraciuk, Andrew Kondrich, Aris Konstantinidis, Kyle Kosic, Gretchen Krueger, Vishal Kuo, Michael Lampe, Ikai Lan, Teddy Lee, Jan Leike, Jade Leung, Daniel Levy, Chak~Ming Li, Rachel Lim, Molly Lin, Stephanie Lin, Mateusz Litwin, Theresa Lopez, Ryan Lowe,
  Patricia Lue, Anna Makanju, Kim Malfacini, Sam Manning, Todor Markov, Yaniv Markovski, Bianca Martin, Katie Mayer, Andrew Mayne, Bob McGrew, Scott~Mayer McKinney, Christine McLeavey, Paul McMillan, Jake McNeil, David Medina, Aalok Mehta, Jacob Menick, Luke Metz, Andrey Mishchenko, Pamela Mishkin, Vinnie Monaco, Evan Morikawa, Daniel Mossing, Tong Mu, Mira Murati, Oleg Murk, David Mély, Ashvin Nair, Reiichiro Nakano, Rajeev Nayak, Arvind Neelakantan, Richard Ngo, Hyeonwoo Noh, Long Ouyang, Cullen O'Keefe, Jakub Pachocki, Alex Paino, Joe Palermo, Ashley Pantuliano, Giambattista Parascandolo, Joel Parish, Emy Parparita, Alex Passos, Mikhail Pavlov, Andrew Peng, Adam Perelman, Filipe de~Avila Belbute~Peres, Michael Petrov, Henrique~Ponde de~Oliveira~Pinto, Michael, Pokorny, Michelle Pokrass, Vitchyr Pong, Tolly Powell, Alethea Power, Boris Power, Elizabeth Proehl, Raul Puri, Alec Radford, Jack Rae, Aditya Ramesh, Cameron Raymond, Francis Real, Kendra Rimbach, Carl Ross, Bob Rotsted, Henri Roussez, Nick Ryder,
  Mario Saltarelli, Ted Sanders, Shibani Santurkar, Girish Sastry, Heather Schmidt, David Schnurr, John Schulman, Daniel Selsam, Kyla Sheppard, Toki Sherbakov, Jessica Shieh, Sarah Shoker, Pranav Shyam, Szymon Sidor, Eric Sigler, Maddie Simens, Jordan Sitkin, Katarina Slama, Ian Sohl, Benjamin Sokolowsky, Yang Song, Natalie Staudacher, Felipe~Petroski Such, Natalie Summers, Ilya Sutskever, Jie Tang, Nikolas Tezak, Madeleine Thompson, Phil Tillet, Amin Tootoonchian, Elizabeth Tseng, Preston Tuggle, Nick Turley, Jerry Tworek, Juan Felipe~Cerón Uribe, Andrea Vallone, Arun Vijayvergiya, Chelsea Voss, Carroll Wainwright, Justin~Jay Wang, Alvin Wang, Ben Wang, Jonathan Ward, Jason Wei, CJ~Weinmann, Akila Welihinda, Peter Welinder, Jiayi Weng, Lilian Weng, Matt Wiethoff, Dave Willner, Clemens Winter, Samuel Wolrich, Hannah Wong, Lauren Workman, Sherwin Wu, Jeff Wu, Michael Wu, Kai Xiao, Tao Xu, Sarah Yoo, Kevin Yu, Qiming Yuan, Wojciech Zaremba, Rowan Zellers, Chong Zhang, Marvin Zhang, Shengjia Zhao, Tianhao
  Zheng, Juntang Zhuang, William Zhuk, and Barret Zoph. 2023.
\newblock \href {https://arxiv.org/abs/2303.08774} {Gpt-4 technical report}.
\newblock \emph{Preprint}, arXiv:2303.08774.

\bibitem[{Ormazabal et~al.(2024)Ormazabal, Zheng, de~Masson~d'Autume, Yogatama, Fu, Ong, Chen, Lamprecht, Pham, Ong, Aleksiev, Li, Henderson, Bain, Artetxe, Relan, Padlewski, Liu, Chen, Phua, Yang, Tay, Wang, Zhu, and Xie}]{ormazabal2024reka}
Aitor Ormazabal, Che Zheng, Cyprien de~Masson~d'Autume, Dani Yogatama, Deyu Fu, Donovan Ong, Eric Chen, Eugenie Lamprecht, Hai Pham, Isaac Ong, Kaloyan Aleksiev, Lei Li, Matthew Henderson, Max Bain, Mikel Artetxe, Nishant Relan, Piotr Padlewski, Qi~Liu, Ren Chen, Samuel Phua, Yazheng Yang, Yi~Tay, Yuqi Wang, Zhongkai Zhu, and Zhihui Xie. 2024.
\newblock \href {https://arxiv.org/abs/2404.12387} {Reka core, flash, and edge: A series of powerful multimodal language models}.
\newblock \emph{Preprint}, arXiv:2404.12387.

\bibitem[{Ouyang et~al.(2022)Ouyang, Wu, Jiang, Almeida, Wainwright, Mishkin, Zhang, Agarwal, Slama, Ray, Schulman, Hilton, Kelton, Miller, Simens, Askell, Welinder, Christiano, Leike, and Lowe}]{rlhf_paper_ouyang}
Long Ouyang, Jeffrey Wu, Xu~Jiang, Diogo Almeida, Carroll Wainwright, Pamela Mishkin, Chong Zhang, Sandhini Agarwal, Katarina Slama, Alex Ray, John Schulman, Jacob Hilton, Fraser Kelton, Luke Miller, Maddie Simens, Amanda Askell, Peter Welinder, Paul~F Christiano, Jan Leike, and Ryan Lowe. 2022.
\newblock \href {https://proceedings.neurips.cc/paper_files/paper/2022/file/b1efde53be364a73914f58805a001731-Paper-Conference.pdf} {Training language models to follow instructions with human feedback}.
\newblock In \emph{Advances in Neural Information Processing Systems}, volume~35, pages 27730--27744. Curran Associates, Inc.

\bibitem[{Prasad et~al.(2008)Prasad, Dinesh, Lee, Miltsakaki, Robaldo, Joshi, and Webber}]{prasad-etal-2008-penn}
Rashmi Prasad, Nikhil Dinesh, Alan Lee, Eleni Miltsakaki, Livio Robaldo, Aravind Joshi, and Bonnie Webber. 2008.
\newblock \href {http://www.lrec-conf.org/proceedings/lrec2008/pdf/754_paper.pdf} {The {P}enn {D}iscourse {T}ree{B}ank 2.0.}
\newblock In \emph{Proceedings of the Sixth International Conference on Language Resources and Evaluation ({LREC}'08)}, Marrakech, Morocco. European Language Resources Association (ELRA).

\bibitem[{Prasad et~al.(2018)Prasad, Webber, and Lee}]{prasad-etal-2018-discourse}
Rashmi Prasad, Bonnie Webber, and Alan Lee. 2018.
\newblock \href {https://aclanthology.org/W18-4710} {Discourse annotation in the {PDTB}: The next generation}.
\newblock In \emph{Proceedings of the 14th Joint {ACL-ISO} Workshop on Interoperable Semantic Annotation}, pages 87--97, Santa Fe, New Mexico, USA. Association for Computational Linguistics.

\bibitem[{Qwen et~al.(2025)Qwen, :, Yang, Yang, Zhang, Hui, Zheng, Yu, Li, Liu, Huang, Wei, Lin, Yang, Tu, Zhang, Yang, Yang, Zhou, Lin, Dang, Lu, Bao, Yang, Yu, Li, Xue, Zhang, Zhu, Men, Lin, Li, Tang, Xia, Ren, Ren, Fan, Su, Zhang, Wan, Liu, Cui, Zhang, and Qiu}]{qwen2025qwen25technicalreport}
Qwen, :, An~Yang, Baosong Yang, Beichen Zhang, Binyuan Hui, Bo~Zheng, Bowen Yu, Chengyuan Li, Dayiheng Liu, Fei Huang, Haoran Wei, Huan Lin, Jian Yang, Jianhong Tu, Jianwei Zhang, Jianxin Yang, Jiaxi Yang, Jingren Zhou, Junyang Lin, Kai Dang, Keming Lu, Keqin Bao, Kexin Yang, Le~Yu, Mei Li, Mingfeng Xue, Pei Zhang, Qin Zhu, Rui Men, Runji Lin, Tianhao Li, Tianyi Tang, Tingyu Xia, Xingzhang Ren, Xuancheng Ren, Yang Fan, Yang Su, Yichang Zhang, Yu~Wan, Yuqiong Liu, Zeyu Cui, Zhenru Zhang, and Zihan Qiu. 2025.
\newblock \href {https://arxiv.org/abs/2412.15115} {Qwen2.5 technical report}.
\newblock \emph{Preprint}, arXiv:2412.15115.

\bibitem[{Radford et~al.(2018)Radford, Narasimhan, Salimans, and Sutskever}]{radford2018improving}
Alec Radford, Karthik Narasimhan, Tim Salimans, and Ilya Sutskever. 2018.
\newblock Improving language understanding by generative pre-training.

\bibitem[{Radford et~al.(2019)Radford, Wu, Child, Luan, Amodei, and Sutskever}]{radford2019language}
Alec Radford, Jeff Wu, Rewon Child, David Luan, Dario Amodei, and Ilya Sutskever. 2019.
\newblock Language models are unsupervised multitask learners.

\bibitem[{Raffel et~al.(2020)Raffel, Shazeer, Roberts, Lee, Narang, Matena, Zhou, Li, and Liu}]{raffel2020exploring_t5}
Colin Raffel, Noam Shazeer, Adam Roberts, Katherine Lee, Sharan Narang, Michael Matena, Yanqi Zhou, Wei Li, and Peter~J Liu. 2020.
\newblock Exploring the limits of transfer learning with a unified text-to-text transformer.
\newblock \emph{The Journal of Machine Learning Research}, 21(1):5485--5551.

\bibitem[{Rajbhandari et~al.(2020)Rajbhandari, Rasley, Ruwase, and He}]{zero_memory_optimization}
Samyam Rajbhandari, Jeff Rasley, Olatunji Ruwase, and Yuxiong He. 2020.
\newblock Zero: memory optimizations toward training trillion parameter models.
\newblock In \emph{Proceedings of the International Conference for High Performance Computing, Networking, Storage and Analysis}, SC '20. IEEE Press.

\bibitem[{Shazeer and Stern(2018)}]{adafactor_shazeer_2018}
Noam Shazeer and Mitchell Stern. 2018.
\newblock \href {https://proceedings.mlr.press/v80/shazeer18a.html} {Adafactor: Adaptive learning rates with sublinear memory cost}.
\newblock In \emph{Proceedings of the 35th International Conference on Machine Learning}, volume~80 of \emph{Proceedings of Machine Learning Research}, pages 4596--4604. PMLR.

\bibitem[{Shi et~al.(2024)Shi, Min, Lomeli, Zhou, Li, Lin, Smith, Zettlemoyer, tau Yih, and Lewis}]{shi2024incontext}
Weijia Shi, Sewon Min, Maria Lomeli, Chunting Zhou, Margaret Li, Xi~Victoria Lin, Noah~A. Smith, Luke Zettlemoyer, Wen tau Yih, and Mike Lewis. 2024.
\newblock \href {https://openreview.net/forum?id=LXVswInHOo} {In-context pretraining: Language modeling beyond document boundaries}.
\newblock In \emph{The Twelfth International Conference on Learning Representations}.

\bibitem[{Socher et~al.(2013)Socher, Perelygin, Wu, Chuang, Manning, Ng, and Potts}]{socher-etal-2013-recursive}
Richard Socher, Alex Perelygin, Jean Wu, Jason Chuang, Christopher~D. Manning, Andrew Ng, and Christopher Potts. 2013.
\newblock \href {https://www.aclweb.org/anthology/D13-1170} {Recursive deep models for semantic compositionality over a sentiment treebank}.
\newblock In \emph{Proceedings of the 2013 Conference on Empirical Methods in Natural Language Processing}, pages 1631--1642, Seattle, Washington, USA. Association for Computational Linguistics.

\bibitem[{Soldaini et~al.(2024)Soldaini, Kinney, Bhagia, Schwenk, Atkinson, Authur, Bogin, Chandu, Dumas, Elazar, Hofmann, Jha, Kumar, Lucy, Lyu, Lambert, Magnusson, Morrison, Muennighoff, Naik, Nam, Peters, Ravichander, Richardson, Shen, Strubell, Subramani, Tafjord, Walsh, Zettlemoyer, Smith, Hajishirzi, Beltagy, Groeneveld, Dodge, and Lo}]{dolma}
Luca Soldaini, Rodney Kinney, Akshita Bhagia, Dustin Schwenk, David Atkinson, Russell Authur, Ben Bogin, Khyathi Chandu, Jennifer Dumas, Yanai Elazar, Valentin Hofmann, Ananya~Harsh Jha, Sachin Kumar, Li~Lucy, Xinxi Lyu, Nathan Lambert, Ian Magnusson, Jacob Morrison, Niklas Muennighoff, Aakanksha Naik, Crystal Nam, Matthew~E. Peters, Abhilasha Ravichander, Kyle Richardson, Zejiang Shen, Emma Strubell, Nishant Subramani, Oyvind Tafjord, Pete Walsh, Luke Zettlemoyer, Noah~A. Smith, Hannaneh Hajishirzi, Iz~Beltagy, Dirk Groeneveld, Jesse Dodge, and Kyle Lo. 2024.
\newblock \href {https://arxiv.org/abs/2402.00159} {{Dolma: An Open Corpus of Three Trillion Tokens for Language Model Pretraining Research}}.
\newblock \emph{arXiv preprint}.

\bibitem[{Sutawika et~al.(2024)Sutawika, Komatsuzaki, and Raffel}]{2024PileT5}
Lintang Sutawika, Aran Komatsuzaki, and Colin Raffel. 2024.
\newblock \href {https://blog.eleuther.ai/pile-t5/} {Pile-t5}.
\newblock Blog post.

\bibitem[{Tay et~al.(2022)Tay, Dehghani, Tran, Garc{\'i}a, Wei, Wang, Chung, Bahri, Schuster, Zheng, Zhou, Houlsby, and Metzler}]{Tay2022UL2UL}
Yi~Tay, Mostafa Dehghani, Vinh~Q. Tran, Xavier Garc{\'i}a, Jason Wei, Xuezhi Wang, Hyung~Won Chung, Dara Bahri, Tal Schuster, Huaixiu~Steven Zheng, Denny Zhou, Neil Houlsby, and Donald Metzler. 2022.
\newblock \href {https://api.semanticscholar.org/CorpusID:252780443} {Ul2: Unifying language learning paradigms}.
\newblock In \emph{International Conference on Learning Representations}.

\bibitem[{Team et~al.(2023)Team, Anil, Borgeaud, Wu, Alayrac, Yu, Soricut, Schalkwyk, Dai, Hauth, Millican, Silver, Petrov, Johnson, Antonoglou, Schrittwieser, Glaese, Chen, Pitler, Lillicrap, Lazaridou, Firat, Molloy, Isard, Barham, Hennigan, Lee, Viola, Reynolds, Xu, Doherty, Collins, Meyer, Rutherford, Moreira, Ayoub, Goel, Tucker, Piqueras, Krikun, Barr, Savinov, Danihelka, Roelofs, White, Andreassen, von Glehn, Yagati, Kazemi, Gonzalez, Khalman, Sygnowski, Frechette, Smith, Culp, Proleev, Luan, Chen, Lottes, Schucher, Lebron, Rrustemi, Clay, Crone, Kocisky, Zhao, Perz, Yu, Howard, Bloniarz, Rae, Lu, Sifre, Maggioni, Alcober, Garrette, Barnes, Thakoor, Austin, Barth-Maron, Wong, Joshi, Chaabouni, Fatiha, Ahuja, Liu, Li, Cogan, Chen, Jia, Gu, Zhang, Grimstad, Hartman, Chadwick, Tomar, Garcia, Senter, Taropa, Pillai, Devlin, Laskin, de~Las~Casas, Valter, Tao, Blanco, Badia, Reitter, Chen, Brennan, Rivera, Brin, Iqbal, Surita, Labanowski, Rao, Winkler, Parisotto, Gu, Olszewska, Zhang, Addanki, Miech, Louis,
  Shafey, Teplyashin, Brown, Catt, Attaluri, Balaguer, Xiang, Wang, Ashwood, Briukhov, Webson, Ganapathy, Sanghavi, Kannan, Chang, Stjerngren, Djolonga, Sun, Bapna, Aitchison, Pejman, Michalewski, Yu, Wang, Love, Ahn, Bloxwich, Han, Humphreys, Sellam, Bradbury, Godbole, Samangooei, Damoc, Kaskasoli, Arnold, Vasudevan, Agrawal, Riesa, Lepikhin, Tanburn, Srinivasan, Lim, Hodkinson, Shyam, Ferret, Hand, Garg, Paine, Li, Li, Giang, Neitz, Abbas, York, Reid, Cole, Chowdhery, Das, Rogozińska, Nikolaev, Sprechmann, Nado, Zilka, Prost, He, Monteiro, Mishra, Welty, Newlan, Jia, Allamanis, Hu, de~Liedekerke, Gilmer, Saroufim, Rijhwani, Hou, Shrivastava, Baddepudi, Goldin, Ozturel, Cassirer, Xu, Sohn, Sachan, Amplayo, Swanson, Petrova, Narayan, Guez, Brahma, Landon, Patel, Zhao, Villela, Wang, Jia, Rahtz, Giménez, Yeung, Lin, Keeling, Georgiev, Mincu, Wu, Haykal, Saputro, Vodrahalli, Qin, Cankara, Sharma, Fernando, Hawkins, Neyshabur, Kim, Hutter, Agrawal, Castro-Ros, van~den Driessche, Wang, Yang, yiin Chang,
  Komarek, McIlroy, Lučić, Zhang, Farhan, Sharman, Natsev, Michel, Cheng, Bansal, Qiao, Cao, Shakeri, Butterfield, Chung, Rubenstein, Agrawal, Mensch, Soparkar, Lenc, Chung, Pope, Maggiore, Kay, Jhakra, Wang, Maynez, Phuong, Tobin, Tacchetti, Trebacz, Robinson, Katariya, Riedel, Bailey, Xiao, Ghelani, Aroyo, Slone, Houlsby, Xiong, Yang, Gribovskaya, Adler, Wirth, Lee, Li, Kagohara, Pavagadhi, Bridgers, Bortsova, Ghemawat, Ahmed, Liu, Powell, Bolina, Iinuma, Zablotskaia, Besley, Chung, Dozat, Comanescu, Si, Greer, Su, Polacek, Kaufman, Tokumine, Hu, Buchatskaya, Miao, Elhawaty, Siddhant, Tomasev, Xing, Greer, Miller, Ashraf, Roy, Zhang, Ma, Filos, Besta, Blevins, Klimenko, Yeh, Changpinyo, Mu, Chang, Pajarskas, Muir, Cohen, Lan, Haridasan, Marathe, Hansen, Douglas, Samuel, Wang, Austin, Lan, Jiang, Chiu, Lorenzo, Sjösund, Cevey, Gleicher, Avrahami, Boral, Srinivasan, Selo, May, Aisopos, Hussenot, Soares, Baumli, Chang, Recasens, Caine, Pritzel, Pavetic, Pardo, Gergely, Frye, Ramasesh, Horgan, Badola,
  Kassner, Roy, Dyer, Campos, Tomala, Tang, Badawy, White, Mustafa, Lang, Jindal, Vikram, Gong, Caelles, Hemsley, Thornton, Feng, Stokowiec, Zheng, Thacker, Çağlar Ünlü, Zhang, Saleh, Svensson, Bileschi, Patil, Anand, Ring, Tsihlas, Vezer, Selvi, Shevlane, Rodriguez, Kwiatkowski, Daruki, Rong, Dafoe, FitzGerald, Gu-Lemberg, Khan, Hendricks, Pellat, Feinberg, Cobon-Kerr, Sainath, Rauh, Hashemi, Ives, Hasson, Li, Noland, Cao, Byrd, Hou, Wang, Sottiaux, Paganini, Lespiau, Moufarek, Hassan, Shivakumar, van Amersfoort, Mandhane, Joshi, Goyal, Tung, Brock, Sheahan, Misra, Li, Rakićević, Dehghani, Liu, Mittal, Oh, Noury, Sezener, Huot, Lamm, Cao, Chen, Elsayed, Chi, Mahdieh, Tenney, Hua, Petrychenko, Kane, Scandinaro, Jain, Uesato, Datta, Sadovsky, Bunyan, Rabiej, Wu, Zhang, Vasudevan, Leurent, Alnahlawi, Georgescu, Wei, Zheng, Chan, Rabinovitch, Stanczyk, Zhang, Steiner, Naskar, Azzam, Johnson, Paszke, Chiu, Elias, Mohiuddin, Muhammad, Miao, Lee, Vieillard, Potluri, Park, Davoodi, Zhang, Stanway, Garmon,
  Karmarkar, Dong, Lee, Kumar, Zhou, Evens, Isaac, Chen, Jia, Levskaya, Zhu, Gorgolewski, Grabowski, Mao, Magni, Yao, Snaider, Casagrande, Suganthan, Palmer, Irving, Loper, Faruqui, Arkatkar, Chen, Shafran, Fink, Castaño, Giannoumis, Kim, Rybiński, Sreevatsa, Prendki, Soergel, Goedeckemeyer, Gierke, Jafari, Gaba, Wiesner, Wright, Wei, Vashisht, Kulizhskaya, Hoover, Le, Li, Iwuanyanwu, Liu, Ramirez, Khorlin, Cui, LIN, Georgiev, Wu, Aguilar, Pallo, Chakladar, Repina, Wu, van~der Weide, Ponnapalli, Kaplan, Simsa, Li, Dousse, Yang, Piper, Ie, Lui, Pasumarthi, Lintz, Vijayakumar, Thiet, Andor, Valenzuela, Paduraru, Peng, Lee, Zhang, Greene, Nguyen, Kurylowicz, Velury, Krause, Hardin, Dixon, Janzer, Choo, Feng, Zhang, Singhal, Latkar, Zhang, Le, Abellan, Du, McKinnon, Antropova, Bolukbasi, Keller, Reid, Finchelstein, Raad, Crocker, Hawkins, Dadashi, Gaffney, Lall, Franko, Filonov, Bulanova, Leblond, Yadav, Chung, Askham, Cobo, Xu, Fischer, Xu, Sorokin, Alberti, Lin, Evans, Zhou, Dimitriev, Forbes, Banarse, Tung,
  Liu, Omernick, Bishop, Kumar, Sterneck, Foley, Jain, Mishra, Xia, Bos, Cideron, Amid, Piccinno, Wang, Banzal, Gurita, Noga, Shah, Mankowitz, Polozov, Kushman, Krakovna, Brown, Bateni, Duan, Firoiu, Thotakuri, Natan, Mohananey, Geist, Mudgal, Girgin, Li, Ye, Roval, Tojo, Kwong, Lee-Thorp, Yew, Yuan, Bagri, Sinopalnikov, Ramos, Mellor, Sharma, Severyn, Lai, Wu, Cheng, Miller, Sonnerat, Vnukov, Greig, Beattie, Caveness, Bai, Eisenschlos, Korchemniy, Tsai, Jasarevic, Kong, Dao, Zheng, Liu, Yang, Zhu, Geller, Teh, Sanmiya, Gladchenko, Trdin, Sozanschi, Toyama, Rosen, Tavakkol, Xue, Elkind, Woodman, Carpenter, Papamakarios, Kemp, Kafle, Grunina, Sinha, Talbert, Goyal, Wu, Owusu-Afriyie, Du, Thornton, Pont-Tuset, Narayana, Li, Fatehi, Wieting, Ajmeri, Uria, Zhu, Ko, Knight, Héliou, Niu, Gu, Pang, Tran, Li, Levine, Stolovich, Kalb, Santamaria-Fernandez, Goenka, Yustalim, Strudel, Elqursh, Lakshminarayanan, Deck, Upadhyay, Lee, Dusenberry, Li, Wang, Levin, Hoffmann, Holtmann-Rice, Bachem, Yue, Arora, Malmi,
  Mirylenka, Tan, Koh, Yeganeh, Põder, Zheng, Pongetti, Tariq, Sun, Ionita, Seyedhosseini, Tafti, Kotikalapudi, Liu, Gulati, Liu, Ye, Chrzaszcz, Wang, Sethi, Li, Brown, Singh, Fan, Parisi, Stanton, Kuang, Koverkathu, Choquette-Choo, Li, Lu, Ittycheriah, Shroff, Sun, Varadarajan, Bahargam, Willoughby, Gaddy, Dasgupta, Desjardins, Cornero, Robenek, Mittal, Albrecht, Shenoy, Moiseev, Jacobsson, Ghaffarkhah, Rivière, Walton, Crepy, Parrish, Liu, Zhou, Farabet, Radebaugh, Srinivasan, van~der Salm, Fidjeland, Scellato, Latorre-Chimoto, Klimczak-Plucińska, Bridson, de~Cesare, Hudson, Mendolicchio, Walker, Morris, Penchev, Mauger, Guseynov, Reid, Odoom, Loher, Cotruta, Yenugula, Grewe, Petrushkina, Duerig, Sanchez, Yadlowsky, Shen, Globerson, Kurzrok, Webb, Dua, Li, Lahoti, Bhupatiraju, Hurt, Qureshi, Agarwal, Shani, Eyal, Khare, Belle, Wang, Tekur, Kale, Wei, Sang, Saeta, Liechty, Sun, Zhao, Lee, Nayak, Fritz, Vuyyuru, Aslanides, Vyas, Wicke, Ma, Bilal, Eltyshev, Balle, Martin, Cate, Manyika, Amiri, Kim, Xiong,
  Kang, Luisier, Tripuraneni, Madras, Guo, Waters, Wang, Ainslie, Baldridge, Zhang, Pruthi, Bauer, Yang, Mansour, Gelman, Xu, Polovets, Liu, Cai, Chen, Sheng, Xue, Ozair, Yu, Angermueller, Li, Wang, Wiesinger, Koukoumidis, Tian, Iyer, Gurumurthy, Goldenson, Shah, Blake, Yu, Urbanowicz, Palomaki, Fernando, Brooks, Durden, Mehta, Momchev, Rahimtoroghi, Georgaki, Raul, Ruder, Redshaw, Lee, Jalan, Li, Perng, Hechtman, Schuh, Nasr, Chen, Milan, Mikulik, Strohman, Franco, Green, Hassabis, Kavukcuoglu, Dean, and Vinyals}]{geminiteam2023gemini}
Gemini Team, Rohan Anil, Sebastian Borgeaud, Yonghui Wu, Jean-Baptiste Alayrac, Jiahui Yu, Radu Soricut, Johan Schalkwyk, Andrew~M. Dai, Anja Hauth, Katie Millican, David Silver, Slav Petrov, Melvin Johnson, Ioannis Antonoglou, Julian Schrittwieser, Amelia Glaese, Jilin Chen, Emily Pitler, Timothy Lillicrap, Angeliki Lazaridou, Orhan Firat, James Molloy, Michael Isard, Paul~R. Barham, Tom Hennigan, Benjamin Lee, Fabio Viola, Malcolm Reynolds, Yuanzhong Xu, Ryan Doherty, Eli Collins, Clemens Meyer, Eliza Rutherford, Erica Moreira, Kareem Ayoub, Megha Goel, George Tucker, Enrique Piqueras, Maxim Krikun, Iain Barr, Nikolay Savinov, Ivo Danihelka, Becca Roelofs, Anaïs White, Anders Andreassen, Tamara von Glehn, Lakshman Yagati, Mehran Kazemi, Lucas Gonzalez, Misha Khalman, Jakub Sygnowski, Alexandre Frechette, Charlotte Smith, Laura Culp, Lev Proleev, Yi~Luan, Xi~Chen, James Lottes, Nathan Schucher, Federico Lebron, Alban Rrustemi, Natalie Clay, Phil Crone, Tomas Kocisky, Jeffrey Zhao, Bartek Perz, Dian Yu,
  Heidi Howard, Adam Bloniarz, Jack~W. Rae, Han Lu, Laurent Sifre, Marcello Maggioni, Fred Alcober, Dan Garrette, Megan Barnes, Shantanu Thakoor, Jacob Austin, Gabriel Barth-Maron, William Wong, Rishabh Joshi, Rahma Chaabouni, Deeni Fatiha, Arun Ahuja, Ruibo Liu, Yunxuan Li, Sarah Cogan, Jeremy Chen, Chao Jia, Chenjie Gu, Qiao Zhang, Jordan Grimstad, Ale~Jakse Hartman, Martin Chadwick, Gaurav~Singh Tomar, Xavier Garcia, Evan Senter, Emanuel Taropa, Thanumalayan~Sankaranarayana Pillai, Jacob Devlin, Michael Laskin, Diego de~Las~Casas, Dasha Valter, Connie Tao, Lorenzo Blanco, Adrià~Puigdomènech Badia, David Reitter, Mianna Chen, Jenny Brennan, Clara Rivera, Sergey Brin, Shariq Iqbal, Gabriela Surita, Jane Labanowski, Abhi Rao, Stephanie Winkler, Emilio Parisotto, Yiming Gu, Kate Olszewska, Yujing Zhang, Ravi Addanki, Antoine Miech, Annie Louis, Laurent~El Shafey, Denis Teplyashin, Geoff Brown, Elliot Catt, Nithya Attaluri, Jan Balaguer, Jackie Xiang, Pidong Wang, Zoe Ashwood, Anton Briukhov, Albert Webson,
  Sanjay Ganapathy, Smit Sanghavi, Ajay Kannan, Ming-Wei Chang, Axel Stjerngren, Josip Djolonga, Yuting Sun, Ankur Bapna, Matthew Aitchison, Pedram Pejman, Henryk Michalewski, Tianhe Yu, Cindy Wang, Juliette Love, Junwhan Ahn, Dawn Bloxwich, Kehang Han, Peter Humphreys, Thibault Sellam, James Bradbury, Varun Godbole, Sina Samangooei, Bogdan Damoc, Alex Kaskasoli, Sébastien M.~R. Arnold, Vijay Vasudevan, Shubham Agrawal, Jason Riesa, Dmitry Lepikhin, Richard Tanburn, Srivatsan Srinivasan, Hyeontaek Lim, Sarah Hodkinson, Pranav Shyam, Johan Ferret, Steven Hand, Ankush Garg, Tom~Le Paine, Jian Li, Yujia Li, Minh Giang, Alexander Neitz, Zaheer Abbas, Sarah York, Machel Reid, Elizabeth Cole, Aakanksha Chowdhery, Dipanjan Das, Dominika Rogozińska, Vitaly Nikolaev, Pablo Sprechmann, Zachary Nado, Lukas Zilka, Flavien Prost, Luheng He, Marianne Monteiro, Gaurav Mishra, Chris Welty, Josh Newlan, Dawei Jia, Miltiadis Allamanis, Clara~Huiyi Hu, Raoul de~Liedekerke, Justin Gilmer, Carl Saroufim, Shruti Rijhwani, Shaobo
  Hou, Disha Shrivastava, Anirudh Baddepudi, Alex Goldin, Adnan Ozturel, Albin Cassirer, Yunhan Xu, Daniel Sohn, Devendra Sachan, Reinald~Kim Amplayo, Craig Swanson, Dessie Petrova, Shashi Narayan, Arthur Guez, Siddhartha Brahma, Jessica Landon, Miteyan Patel, Ruizhe Zhao, Kevin Villela, Luyu Wang, Wenhao Jia, Matthew Rahtz, Mai Giménez, Legg Yeung, Hanzhao Lin, James Keeling, Petko Georgiev, Diana Mincu, Boxi Wu, Salem Haykal, Rachel Saputro, Kiran Vodrahalli, James Qin, Zeynep Cankara, Abhanshu Sharma, Nick Fernando, Will Hawkins, Behnam Neyshabur, Solomon Kim, Adrian Hutter, Priyanka Agrawal, Alex Castro-Ros, George van~den Driessche, Tao Wang, Fan Yang, Shuo yiin Chang, Paul Komarek, Ross McIlroy, Mario Lučić, Guodong Zhang, Wael Farhan, Michael Sharman, Paul Natsev, Paul Michel, Yong Cheng, Yamini Bansal, Siyuan Qiao, Kris Cao, Siamak Shakeri, Christina Butterfield, Justin Chung, Paul~Kishan Rubenstein, Shivani Agrawal, Arthur Mensch, Kedar Soparkar, Karel Lenc, Timothy Chung, Aedan Pope, Loren
  Maggiore, Jackie Kay, Priya Jhakra, Shibo Wang, Joshua Maynez, Mary Phuong, Taylor Tobin, Andrea Tacchetti, Maja Trebacz, Kevin Robinson, Yash Katariya, Sebastian Riedel, Paige Bailey, Kefan Xiao, Nimesh Ghelani, Lora Aroyo, Ambrose Slone, Neil Houlsby, Xuehan Xiong, Zhen Yang, Elena Gribovskaya, Jonas Adler, Mateo Wirth, Lisa Lee, Music Li, Thais Kagohara, Jay Pavagadhi, Sophie Bridgers, Anna Bortsova, Sanjay Ghemawat, Zafarali Ahmed, Tianqi Liu, Richard Powell, Vijay Bolina, Mariko Iinuma, Polina Zablotskaia, James Besley, Da-Woon Chung, Timothy Dozat, Ramona Comanescu, Xiance Si, Jeremy Greer, Guolong Su, Martin Polacek, Raphaël~Lopez Kaufman, Simon Tokumine, Hexiang Hu, Elena Buchatskaya, Yingjie Miao, Mohamed Elhawaty, Aditya Siddhant, Nenad Tomasev, Jinwei Xing, Christina Greer, Helen Miller, Shereen Ashraf, Aurko Roy, Zizhao Zhang, Ada Ma, Angelos Filos, Milos Besta, Rory Blevins, Ted Klimenko, Chih-Kuan Yeh, Soravit Changpinyo, Jiaqi Mu, Oscar Chang, Mantas Pajarskas, Carrie Muir, Vered Cohen,
  Charline~Le Lan, Krishna Haridasan, Amit Marathe, Steven Hansen, Sholto Douglas, Rajkumar Samuel, Mingqiu Wang, Sophia Austin, Chang Lan, Jiepu Jiang, Justin Chiu, Jaime~Alonso Lorenzo, Lars~Lowe Sjösund, Sébastien Cevey, Zach Gleicher, Thi Avrahami, Anudhyan Boral, Hansa Srinivasan, Vittorio Selo, Rhys May, Konstantinos Aisopos, Léonard Hussenot, Livio~Baldini Soares, Kate Baumli, Michael~B. Chang, Adrià Recasens, Ben Caine, Alexander Pritzel, Filip Pavetic, Fabio Pardo, Anita Gergely, Justin Frye, Vinay Ramasesh, Dan Horgan, Kartikeya Badola, Nora Kassner, Subhrajit Roy, Ethan Dyer, Víctor Campos, Alex Tomala, Yunhao Tang, Dalia~El Badawy, Elspeth White, Basil Mustafa, Oran Lang, Abhishek Jindal, Sharad Vikram, Zhitao Gong, Sergi Caelles, Ross Hemsley, Gregory Thornton, Fangxiaoyu Feng, Wojciech Stokowiec, Ce~Zheng, Phoebe Thacker, Çağlar Ünlü, Zhishuai Zhang, Mohammad Saleh, James Svensson, Max Bileschi, Piyush Patil, Ankesh Anand, Roman Ring, Katerina Tsihlas, Arpi Vezer, Marco Selvi, Toby
  Shevlane, Mikel Rodriguez, Tom Kwiatkowski, Samira Daruki, Keran Rong, Allan Dafoe, Nicholas FitzGerald, Keren Gu-Lemberg, Mina Khan, Lisa~Anne Hendricks, Marie Pellat, Vladimir Feinberg, James Cobon-Kerr, Tara Sainath, Maribeth Rauh, Sayed~Hadi Hashemi, Richard Ives, Yana Hasson, YaGuang Li, Eric Noland, Yuan Cao, Nathan Byrd, Le~Hou, Qingze Wang, Thibault Sottiaux, Michela Paganini, Jean-Baptiste Lespiau, Alexandre Moufarek, Samer Hassan, Kaushik Shivakumar, Joost van Amersfoort, Amol Mandhane, Pratik Joshi, Anirudh Goyal, Matthew Tung, Andrew Brock, Hannah Sheahan, Vedant Misra, Cheng Li, Nemanja Rakićević, Mostafa Dehghani, Fangyu Liu, Sid Mittal, Junhyuk Oh, Seb Noury, Eren Sezener, Fantine Huot, Matthew Lamm, Nicola~De Cao, Charlie Chen, Gamaleldin Elsayed, Ed~Chi, Mahdis Mahdieh, Ian Tenney, Nan Hua, Ivan Petrychenko, Patrick Kane, Dylan Scandinaro, Rishub Jain, Jonathan Uesato, Romina Datta, Adam Sadovsky, Oskar Bunyan, Dominik Rabiej, Shimu Wu, John Zhang, Gautam Vasudevan, Edouard Leurent,
  Mahmoud Alnahlawi, Ionut Georgescu, Nan Wei, Ivy Zheng, Betty Chan, Pam~G Rabinovitch, Piotr Stanczyk, Ye~Zhang, David Steiner, Subhajit Naskar, Michael Azzam, Matthew Johnson, Adam Paszke, Chung-Cheng Chiu, Jaume~Sanchez Elias, Afroz Mohiuddin, Faizan Muhammad, Jin Miao, Andrew Lee, Nino Vieillard, Sahitya Potluri, Jane Park, Elnaz Davoodi, Jiageng Zhang, Jeff Stanway, Drew Garmon, Abhijit Karmarkar, Zhe Dong, Jong Lee, Aviral Kumar, Luowei Zhou, Jonathan Evens, William Isaac, Zhe Chen, Johnson Jia, Anselm Levskaya, Zhenkai Zhu, Chris Gorgolewski, Peter Grabowski, Yu~Mao, Alberto Magni, Kaisheng Yao, Javier Snaider, Norman Casagrande, Paul Suganthan, Evan Palmer, Geoffrey Irving, Edward Loper, Manaal Faruqui, Isha Arkatkar, Nanxin Chen, Izhak Shafran, Michael Fink, Alfonso Castaño, Irene Giannoumis, Wooyeol Kim, Mikołaj Rybiński, Ashwin Sreevatsa, Jennifer Prendki, David Soergel, Adrian Goedeckemeyer, Willi Gierke, Mohsen Jafari, Meenu Gaba, Jeremy Wiesner, Diana~Gage Wright, Yawen Wei, Harsha Vashisht,
  Yana Kulizhskaya, Jay Hoover, Maigo Le, Lu~Li, Chimezie Iwuanyanwu, Lu~Liu, Kevin Ramirez, Andrey Khorlin, Albert Cui, Tian LIN, Marin Georgiev, Marcus Wu, Ricardo Aguilar, Keith Pallo, Abhishek Chakladar, Alena Repina, Xihui Wu, Tom van~der Weide, Priya Ponnapalli, Caroline Kaplan, Jiri Simsa, Shuangfeng Li, Olivier Dousse, Fan Yang, Jeff Piper, Nathan Ie, Minnie Lui, Rama Pasumarthi, Nathan Lintz, Anitha Vijayakumar, Lam~Nguyen Thiet, Daniel Andor, Pedro Valenzuela, Cosmin Paduraru, Daiyi Peng, Katherine Lee, Shuyuan Zhang, Somer Greene, Duc~Dung Nguyen, Paula Kurylowicz, Sarmishta Velury, Sebastian Krause, Cassidy Hardin, Lucas Dixon, Lili Janzer, Kiam Choo, Ziqiang Feng, Biao Zhang, Achintya Singhal, Tejasi Latkar, Mingyang Zhang, Quoc Le, Elena~Allica Abellan, Dayou Du, Dan McKinnon, Natasha Antropova, Tolga Bolukbasi, Orgad Keller, David Reid, Daniel Finchelstein, Maria~Abi Raad, Remi Crocker, Peter Hawkins, Robert Dadashi, Colin Gaffney, Sid Lall, Ken Franko, Egor Filonov, Anna Bulanova, Rémi
  Leblond, Vikas Yadav, Shirley Chung, Harry Askham, Luis~C. Cobo, Kelvin Xu, Felix Fischer, Jun Xu, Christina Sorokin, Chris Alberti, Chu-Cheng Lin, Colin Evans, Hao Zhou, Alek Dimitriev, Hannah Forbes, Dylan Banarse, Zora Tung, Jeremiah Liu, Mark Omernick, Colton Bishop, Chintu Kumar, Rachel Sterneck, Ryan Foley, Rohan Jain, Swaroop Mishra, Jiawei Xia, Taylor Bos, Geoffrey Cideron, Ehsan Amid, Francesco Piccinno, Xingyu Wang, Praseem Banzal, Petru Gurita, Hila Noga, Premal Shah, Daniel~J. Mankowitz, Alex Polozov, Nate Kushman, Victoria Krakovna, Sasha Brown, MohammadHossein Bateni, Dennis Duan, Vlad Firoiu, Meghana Thotakuri, Tom Natan, Anhad Mohananey, Matthieu Geist, Sidharth Mudgal, Sertan Girgin, Hui Li, Jiayu Ye, Ofir Roval, Reiko Tojo, Michael Kwong, James Lee-Thorp, Christopher Yew, Quan Yuan, Sumit Bagri, Danila Sinopalnikov, Sabela Ramos, John Mellor, Abhishek Sharma, Aliaksei Severyn, Jonathan Lai, Kathy Wu, Heng-Tze Cheng, David Miller, Nicolas Sonnerat, Denis Vnukov, Rory Greig, Jennifer
  Beattie, Emily Caveness, Libin Bai, Julian Eisenschlos, Alex Korchemniy, Tomy Tsai, Mimi Jasarevic, Weize Kong, Phuong Dao, Zeyu Zheng, Frederick Liu, Fan Yang, Rui Zhu, Mark Geller, Tian~Huey Teh, Jason Sanmiya, Evgeny Gladchenko, Nejc Trdin, Andrei Sozanschi, Daniel Toyama, Evan Rosen, Sasan Tavakkol, Linting Xue, Chen Elkind, Oliver Woodman, John Carpenter, George Papamakarios, Rupert Kemp, Sushant Kafle, Tanya Grunina, Rishika Sinha, Alice Talbert, Abhimanyu Goyal, Diane Wu, Denese Owusu-Afriyie, Cosmo Du, Chloe Thornton, Jordi Pont-Tuset, Pradyumna Narayana, Jing Li, Sabaer Fatehi, John Wieting, Omar Ajmeri, Benigno Uria, Tao Zhu, Yeongil Ko, Laura Knight, Amélie Héliou, Ning Niu, Shane Gu, Chenxi Pang, Dustin Tran, Yeqing Li, Nir Levine, Ariel Stolovich, Norbert Kalb, Rebeca Santamaria-Fernandez, Sonam Goenka, Wenny Yustalim, Robin Strudel, Ali Elqursh, Balaji Lakshminarayanan, Charlie Deck, Shyam Upadhyay, Hyo Lee, Mike Dusenberry, Zonglin Li, Xuezhi Wang, Kyle Levin, Raphael Hoffmann, Dan
  Holtmann-Rice, Olivier Bachem, Summer Yue, Sho Arora, Eric Malmi, Daniil Mirylenka, Qijun Tan, Christy Koh, Soheil~Hassas Yeganeh, Siim Põder, Steven Zheng, Francesco Pongetti, Mukarram Tariq, Yanhua Sun, Lucian Ionita, Mojtaba Seyedhosseini, Pouya Tafti, Ragha Kotikalapudi, Zhiyu Liu, Anmol Gulati, Jasmine Liu, Xinyu Ye, Bart Chrzaszcz, Lily Wang, Nikhil Sethi, Tianrun Li, Ben Brown, Shreya Singh, Wei Fan, Aaron Parisi, Joe Stanton, Chenkai Kuang, Vinod Koverkathu, Christopher~A. Choquette-Choo, Yunjie Li, TJ~Lu, Abe Ittycheriah, Prakash Shroff, Pei Sun, Mani Varadarajan, Sanaz Bahargam, Rob Willoughby, David Gaddy, Ishita Dasgupta, Guillaume Desjardins, Marco Cornero, Brona Robenek, Bhavishya Mittal, Ben Albrecht, Ashish Shenoy, Fedor Moiseev, Henrik Jacobsson, Alireza Ghaffarkhah, Morgane Rivière, Alanna Walton, Clément Crepy, Alicia Parrish, Yuan Liu, Zongwei Zhou, Clement Farabet, Carey Radebaugh, Praveen Srinivasan, Claudia van~der Salm, Andreas Fidjeland, Salvatore Scellato, Eri Latorre-Chimoto,
  Hanna Klimczak-Plucińska, David Bridson, Dario de~Cesare, Tom Hudson, Piermaria Mendolicchio, Lexi Walker, Alex Morris, Ivo Penchev, Matthew Mauger, Alexey Guseynov, Alison Reid, Seth Odoom, Lucia Loher, Victor Cotruta, Madhavi Yenugula, Dominik Grewe, Anastasia Petrushkina, Tom Duerig, Antonio Sanchez, Steve Yadlowsky, Amy Shen, Amir Globerson, Adam Kurzrok, Lynette Webb, Sahil Dua, Dong Li, Preethi Lahoti, Surya Bhupatiraju, Dan Hurt, Haroon Qureshi, Ananth Agarwal, Tomer Shani, Matan Eyal, Anuj Khare, Shreyas~Rammohan Belle, Lei Wang, Chetan Tekur, Mihir~Sanjay Kale, Jinliang Wei, Ruoxin Sang, Brennan Saeta, Tyler Liechty, Yi~Sun, Yao Zhao, Stephan Lee, Pandu Nayak, Doug Fritz, Manish~Reddy Vuyyuru, John Aslanides, Nidhi Vyas, Martin Wicke, Xiao Ma, Taylan Bilal, Evgenii Eltyshev, Daniel Balle, Nina Martin, Hardie Cate, James Manyika, Keyvan Amiri, Yelin Kim, Xi~Xiong, Kai Kang, Florian Luisier, Nilesh Tripuraneni, David Madras, Mandy Guo, Austin Waters, Oliver Wang, Joshua Ainslie, Jason Baldridge, Han
  Zhang, Garima Pruthi, Jakob Bauer, Feng Yang, Riham Mansour, Jason Gelman, Yang Xu, George Polovets, Ji~Liu, Honglong Cai, Warren Chen, XiangHai Sheng, Emily Xue, Sherjil Ozair, Adams Yu, Christof Angermueller, Xiaowei Li, Weiren Wang, Julia Wiesinger, Emmanouil Koukoumidis, Yuan Tian, Anand Iyer, Madhu Gurumurthy, Mark Goldenson, Parashar Shah, MK~Blake, Hongkun Yu, Anthony Urbanowicz, Jennimaria Palomaki, Chrisantha Fernando, Kevin Brooks, Ken Durden, Harsh Mehta, Nikola Momchev, Elahe Rahimtoroghi, Maria Georgaki, Amit Raul, Sebastian Ruder, Morgan Redshaw, Jinhyuk Lee, Komal Jalan, Dinghua Li, Ginger Perng, Blake Hechtman, Parker Schuh, Milad Nasr, Mia Chen, Kieran Milan, Vladimir Mikulik, Trevor Strohman, Juliana Franco, Tim Green, Demis Hassabis, Koray Kavukcuoglu, Jeffrey Dean, and Oriol Vinyals. 2023.
\newblock \href {https://arxiv.org/abs/2312.11805} {Gemini: A family of highly capable multimodal models}.
\newblock \emph{Preprint}, arXiv:2312.11805.

\bibitem[{Touvron et~al.(2023)Touvron, Lavril, Izacard, Martinet, Lachaux, Lacroix, Rozière, Goyal, Hambro, Azhar, Rodriguez, Joulin, Grave, and Lample}]{touvron2023llama}
Hugo Touvron, Thibaut Lavril, Gautier Izacard, Xavier Martinet, Marie-Anne Lachaux, Timothée Lacroix, Baptiste Rozière, Naman Goyal, Eric Hambro, Faisal Azhar, Aurelien Rodriguez, Armand Joulin, Edouard Grave, and Guillaume Lample. 2023.
\newblock \href {https://arxiv.org/abs/2302.13971} {Llama: Open and efficient foundation language models}.
\newblock \emph{Preprint}, arXiv:2302.13971.

\bibitem[{Wang et~al.(2018)Wang, Singh, Michael, Hill, Levy, and Bowman}]{wang-etal-2018-glue}
Alex Wang, Amanpreet Singh, Julian Michael, Felix Hill, Omer Levy, and Samuel Bowman. 2018.
\newblock \href {https://doi.org/10.18653/v1/W18-5446} {{GLUE}: A multi-task benchmark and analysis platform for natural language understanding}.
\newblock In \emph{Proceedings of the 2018 {EMNLP} Workshop {B}lackbox{NLP}: Analyzing and Interpreting Neural Networks for {NLP}}, pages 353--355, Brussels, Belgium. Association for Computational Linguistics.

\bibitem[{Wang et~al.(2023)Wang, Du, Liu, Cai, Yu, Jiang, Wang, Cui, Shi, and Tu}]{wang2023discobenchdiscourseawareevaluationbenchmark}
Longyue Wang, Zefeng Du, Donghuai Liu, Deng Cai, Dian Yu, Haiyun Jiang, Yan Wang, Leyang Cui, Shuming Shi, and Zhaopeng Tu. 2023.
\newblock \href {https://arxiv.org/abs/2307.08074} {Disco-bench: A discourse-aware evaluation benchmark for language modelling}.
\newblock \emph{Preprint}, arXiv:2307.08074.

\bibitem[{Warstadt et~al.(2019)Warstadt, Singh, and Bowman}]{warstadt2019neural}
Alex Warstadt, Amanpreet Singh, and Samuel~R Bowman. 2019.
\newblock Neural network acceptability judgments.
\newblock \emph{Transactions of the Association for Computational Linguistics}, 7:625--641.

\bibitem[{Wei et~al.(2022{\natexlab{a}})Wei, Bosma, Zhao, Guu, Yu, Lester, Du, Dai, and Le}]{wei2022finetuned}
Jason Wei, Maarten Bosma, Vincent Zhao, Kelvin Guu, Adams~Wei Yu, Brian Lester, Nan Du, Andrew~M. Dai, and Quoc~V Le. 2022{\natexlab{a}}.
\newblock \href {https://openreview.net/forum?id=gEZrGCozdqR} {Finetuned language models are zero-shot learners}.
\newblock In \emph{International Conference on Learning Representations}.

\bibitem[{Wei et~al.(2022{\natexlab{b}})Wei, Tay, Bommasani, Raffel, Zoph, Borgeaud, Yogatama, Bosma, Zhou, Metzler, Chi, Hashimoto, Vinyals, Liang, Dean, and Fedus}]{wei2022emergent}
Jason Wei, Yi~Tay, Rishi Bommasani, Colin Raffel, Barret Zoph, Sebastian Borgeaud, Dani Yogatama, Maarten Bosma, Denny Zhou, Donald Metzler, Ed~H. Chi, Tatsunori Hashimoto, Oriol Vinyals, Percy Liang, Jeff Dean, and William Fedus. 2022{\natexlab{b}}.
\newblock \href {https://openreview.net/forum?id=yzkSU5zdwD} {Emergent abilities of large language models}.
\newblock \emph{Transactions on Machine Learning Research}.
\newblock Survey Certification.

\bibitem[{Williams et~al.(2018)Williams, Nangia, and Bowman}]{williams-etal-2018-broad}
Adina Williams, Nikita Nangia, and Samuel Bowman. 2018.
\newblock \href {https://doi.org/10.18653/v1/N18-1101} {A broad-coverage challenge corpus for sentence understanding through inference}.
\newblock In \emph{Proceedings of the 2018 Conference of the North {A}merican Chapter of the Association for Computational Linguistics: Human Language Technologies, Volume 1 (Long Papers)}, pages 1112--1122, New Orleans, Louisiana. Association for Computational Linguistics.

\bibitem[{Wolf et~al.(2019)Wolf, Debut, Sanh, Chaumond, Delangue, Moi, Cistac, Rault, Louf, Funtowicz et~al.}]{wolf2019huggingface}
Thomas Wolf, Lysandre Debut, Victor Sanh, Julien Chaumond, Clement Delangue, Anthony Moi, Pierric Cistac, Tim Rault, R{\'e}mi Louf, Morgan Funtowicz, et~al. 2019.
\newblock Huggingface's transformers: State-of-the-art natural language processing.
\newblock \emph{arXiv preprint arXiv:1910.03771}.

\bibitem[{Wu et~al.(2021)Wu, Wu, Qi, and Huang}]{wu-etal-2021-hi}
Chuhan Wu, Fangzhao Wu, Tao Qi, and Yongfeng Huang. 2021.
\newblock \href {https://doi.org/10.18653/v1/2021.acl-short.107} {Hi-transformer: Hierarchical interactive transformer for efficient and effective long document modeling}.
\newblock In \emph{Proceedings of the 59th Annual Meeting of the Association for Computational Linguistics and the 11th International Joint Conference on Natural Language Processing (Volume 2: Short Papers)}, pages 848--853, Online. Association for Computational Linguistics.

\bibitem[{Wu et~al.(2023)Wu, Kim, Watanabe, Han, McDonald, Weinberger, and Artzi}]{wu2023wav2seq}
Felix Wu, Kwangyoun Kim, Shinji Watanabe, Kyu~J Han, Ryan McDonald, Kilian~Q Weinberger, and Yoav Artzi. 2023.
\newblock Wav2seq: Pre-training speech-to-text encoder-decoder models using pseudo languages.
\newblock In \emph{ICASSP 2023-2023 IEEE International Conference on Acoustics, Speech and Signal Processing (ICASSP)}, pages 1--5. IEEE.

\bibitem[{Xue et~al.(2021)Xue, Constant, Roberts, Kale, Al-Rfou, Siddhant, Barua, and Raffel}]{xue-etal-2021-mt5}
Linting Xue, Noah Constant, Adam Roberts, Mihir Kale, Rami Al-Rfou, Aditya Siddhant, Aditya Barua, and Colin Raffel. 2021.
\newblock \href {https://doi.org/10.18653/v1/2021.naacl-main.41} {m{T}5: A massively multilingual pre-trained text-to-text transformer}.
\newblock In \emph{Proceedings of the 2021 Conference of the North American Chapter of the Association for Computational Linguistics: Human Language Technologies}, pages 483--498, Online. Association for Computational Linguistics.

\bibitem[{Yang et~al.(2021)Yang, Yang, Cer, Law, and Darve}]{yang-etal-2021-universal}
Ziyi Yang, Yinfei Yang, Daniel Cer, Jax Law, and Eric Darve. 2021.
\newblock \href {https://doi.org/10.18653/v1/2021.emnlp-main.502} {Universal sentence representation learning with conditional masked language model}.
\newblock In \emph{Proceedings of the 2021 Conference on Empirical Methods in Natural Language Processing}, pages 6216--6228, Online and Punta Cana, Dominican Republic. Association for Computational Linguistics.

\bibitem[{Yu et~al.(2024)Yu, Gu, Huang, and Li}]{yu_etal_predicting_the_next_sentence}
Shaoyun Yu, Chanyuan Gu, Kexin Huang, and Ping Li. 2024.
\newblock \href {https://doi.org/10.1126/sciadv.adn7744} {Predicting the next sentence (not word) in large language models: What model-brain alignment tells us about discourse comprehension}.
\newblock \emph{Science Advances}, 10(21):eadn7744.

\bibitem[{Zhai et~al.(2022)Zhai, Kolesnikov, Houlsby, and Beyer}]{scaling_vision_transformers}
Xiaohua Zhai, Alexander Kolesnikov, Neil Houlsby, and Lucas Beyer. 2022.
\newblock \href {https://arxiv.org/abs/2106.04560} {Scaling vision transformers}.
\newblock In \emph{CVPR}.

\bibitem[{Zhang et~al.(2020)Zhang, Zhao, Saleh, and Liu}]{zhang2020pegasus}
Jingqing Zhang, Yao Zhao, Mohammad Saleh, and Peter Liu. 2020.
\newblock Pegasus: Pre-training with extracted gap-sentences for abstractive summarization.
\newblock In \emph{International Conference on Machine Learning}, pages 11328--11339. PMLR.

\bibitem[{Zhao et~al.(2022)Zhao, Huang, Chowdhury, Chandrasekaran, McKeown, and Chaturvedi}]{Zhao2022ReadTN_summarization}
Chao Zhao, Tenghao Huang, Somnath Basu~Roy Chowdhury, Muthu~Kumar Chandrasekaran, Kathleen McKeown, and Snigdha Chaturvedi. 2022.
\newblock \href {https://api.semanticscholar.org/CorpusID:247593888} {Read top news first: A document reordering approach for multi-document news summarization}.
\newblock \emph{ArXiv}, abs/2203.10254.

\bibitem[{Zheng et~al.(2023)Zheng, Chiang, Sheng, Li, Zhuang, Wu, Zhuang, Li, Lin, Xing et~al.}]{zheng2023lmsys}
Lianmin Zheng, Wei-Lin Chiang, Ying Sheng, Tianle Li, Siyuan Zhuang, Zhanghao Wu, Yonghao Zhuang, Zhuohan Li, Zi~Lin, Eric Xing, et~al. 2023.
\newblock Lmsys-chat-1m: A large-scale real-world llm conversation dataset.
\newblock \emph{arXiv preprint arXiv:2309.11998}.

\bibitem[{Üstün et~al.(2024)Üstün, Aryabumi, Yong, Ko, D'souza, Onilude, Bhandari, Singh, Ooi, Kayid, Vargus, Blunsom, Longpre, Muennighoff, Fadaee, Kreutzer, and Hooker}]{cohere2024aya}
Ahmet Üstün, Viraat Aryabumi, Zheng-Xin Yong, Wei-Yin Ko, Daniel D'souza, Gbemileke Onilude, Neel Bhandari, Shivalika Singh, Hui-Lee Ooi, Amr Kayid, Freddie Vargus, Phil Blunsom, Shayne Longpre, Niklas Muennighoff, Marzieh Fadaee, Julia Kreutzer, and Sara Hooker. 2024.
\newblock \href {https://arxiv.org/abs/2402.07827} {Aya model: An instruction finetuned open-access multilingual language model}.
\newblock \emph{Preprint}, arXiv:2402.07827.

\end{thebibliography}

\newpage

\appendix

\section{Challenges in replicating T5}
\label{sec:replicating_t5}

\subsection{Our speculations}

We recognize that both DEPTH and T5 CPT models quickly reach a significantly lower reconstruction loss than their FS counterparts. While the original T5's reached a final training loss of $\approx0.75$ (with 512k steps of batch size 128 with packing) \cite{raffel2020exploring_t5}, our training loss reached $\approx0.8$ (after 800k steps of batch size 200 without packing). We propose the following ideas to explain this gap: 

\begin{enumerate}
    \item The baseline model from which CPT models are initialized is trained on 1T tokens. This is $\approx20\times$ greater than the amount of tokens we used to train FS models. 
    \item Our examples never consist of the later text in long documents. By truncating text after 512 tokens, our FS models might miss out on valuable text to train on. 
    \item Our training examples consists of $\approx2\times$ fewer tokens (on average) than examples the original T5 was trained on. Furthermore, there is much greater variance in example lengths in our pre-training experiments. These factors may impact our models' learning dynamics (e.g., learning effective positional representations given the presence of irregular padding patterns).
    \item The T5 baseline plot was reported using a masking probability of $0.15$, which is $2\times$ lower than the one we used. Our higher masking probability makes the reconstruction task more challenging.
    \item T5 models using the T5x framework use aggressively high learning rates, which can lead to a different, and perhaps more effective training dynamics than the ones we found in our FS experiments. Using such high learning rates in our settings caused our models to diverge.  
\end{enumerate}

\subsection{Example packing}
\label{sec:example_packing}

\begin{figure}[h!]
    \centering
    \includegraphics[width=\columnwidth]{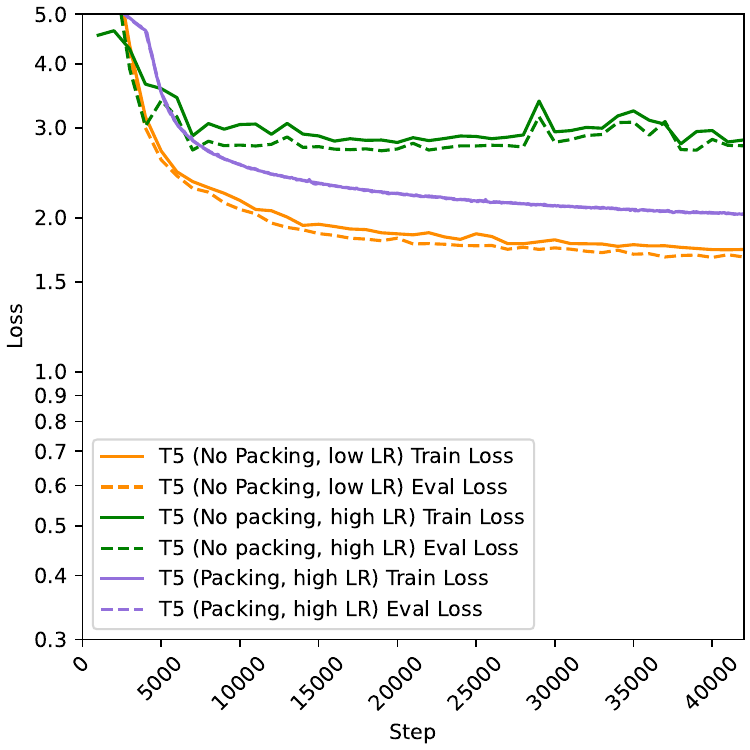}
    \caption{Exploration of packing and learning rates when pre-training T5 models. ``High LR'' corresponds to a learning rate of $1e^{-2}$, while ``Low LR'' corresponds to a learning rate of $1e^{-4}$.}
    \label{fig:recreating_t5}
\end{figure}

Given our choice of avoiding example-packing, we found that we were not able to pre-train T5 with the same hyper-parameters used in \cite{raffel2020exploring_t5}. Specifically, we found that with a batch size of 128 and a learning rate of $1e^{-2}$, our model consistently diverged. This issue persisted with a learning rate of $1e^{-3}$. To stabilize our loss given the absence of packing, we used a lower maximum learning rate ($1e^{-4}$), which is in line with those used to pre-train BERT, SLM, and PMI. On the other hand, we see that given the same training parameters (i.e., learning rate and batch size), pre-training with packing can converge at a high learning rate (see Figure \ref{fig:recreating_t5}).

We speculate that packing acts in a similar way to increasing the batch size during training. The model is exposed to loss on a greater amount of tokens in each optimization step, and is therefore able to generalize even with a larger learning rate.

One possible side effect of avoiding example-packing is the truncation of long examples (examples are not dynamically chunked, so every token past the context limit of 512 is ignored). We empirically find that while T5 suffers greatly from no packing, DEPTH is able to train effectively despite these limitations.

\newpage

\section{DEPTH loss decomposition}
\label{sec:depth_loss_decomposition}

\begin{figure}[h!]
    \begin{minipage}{0.48\textwidth}
        \centering
        \includegraphics[width=\textwidth]{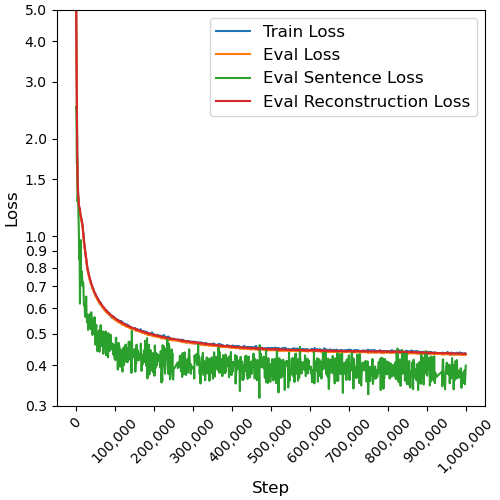}
        \caption{Decomposition of from-scratch pre-training losses (validation) for DEPTH.}
        \label{fig:from_scratch_depth_reconstruction_losses}
    \end{minipage}\hfill
    \begin{minipage}{0.48\textwidth}
        \centering
        \includegraphics[width=\textwidth]{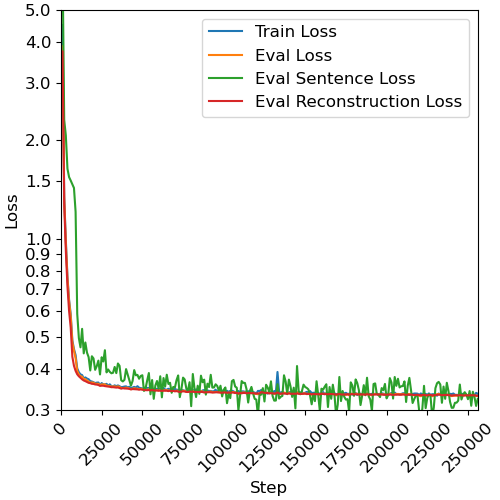}
        \caption{Decomposition of continuous pre-training losses (validation) for DEPTH}
        \label{fig:continuous_depth_validation_losses}
    \end{minipage}
\end{figure}

When we decompose DEPTH's losses in the FS setting, we find that sentence loss is consistently lower than reconstruction loss, but plateus early on into training. The overall loss is dominated by the reconstruction loss, reflected by overlapping lines in Figure \ref{fig:from_scratch_depth_reconstruction_losses}. In the CPT setting (Figure \ref{fig:continuous_depth_validation_losses}), we find that both of DEPTH's losses plateu sooner, and that the sentence loss is approximately equal (though with higher fluctuations) to the reconstruction loss.

We note that DEPTH's loss over sentence tokens in the FC setting, is close to that which the DEPTH-CPT achieved (both in the range of 0.3-0.4, where the more examples are shuffled, the higher the sentence loss). In practice, given comperable ratios of shuffling sentences, CPT DEPTH outperforms FS DEPTH in predicting the next sentence accurately during pre-training ($\approx1\%-3\%$ higher given a fixed shuffling ratio). We speculate that the better representations for non-sentence tokens in the CPT setting is the reason for this performance boost. 

\subsection{Weight of sentence loss}
\label{sec:sentence_token_weight}

We explored the impact of increasing the weight of the sentence loss during DEPTH pre-training. Our ``Baseline'' run is the DEPTH model we reported on in the main body of the paper. In our ``Sentence Weight 1x'' run, our loss is composed of the average loss over sentence tokens plus the average loss over non-sentence tokens (as opposed to the average loss over all tokens). This formulation places increased weight on sentence tokens, since there are significantly fewer sentence tokens than non-sentence tokens. In the ``Sentence Weight 5x'' run, we weighed the sentence loss $5\times$ more than reconstruction loss.

\begin{figure}[h!]
    \centering
    \makebox[\linewidth]{%
        \includegraphics[width=1.5\columnwidth]{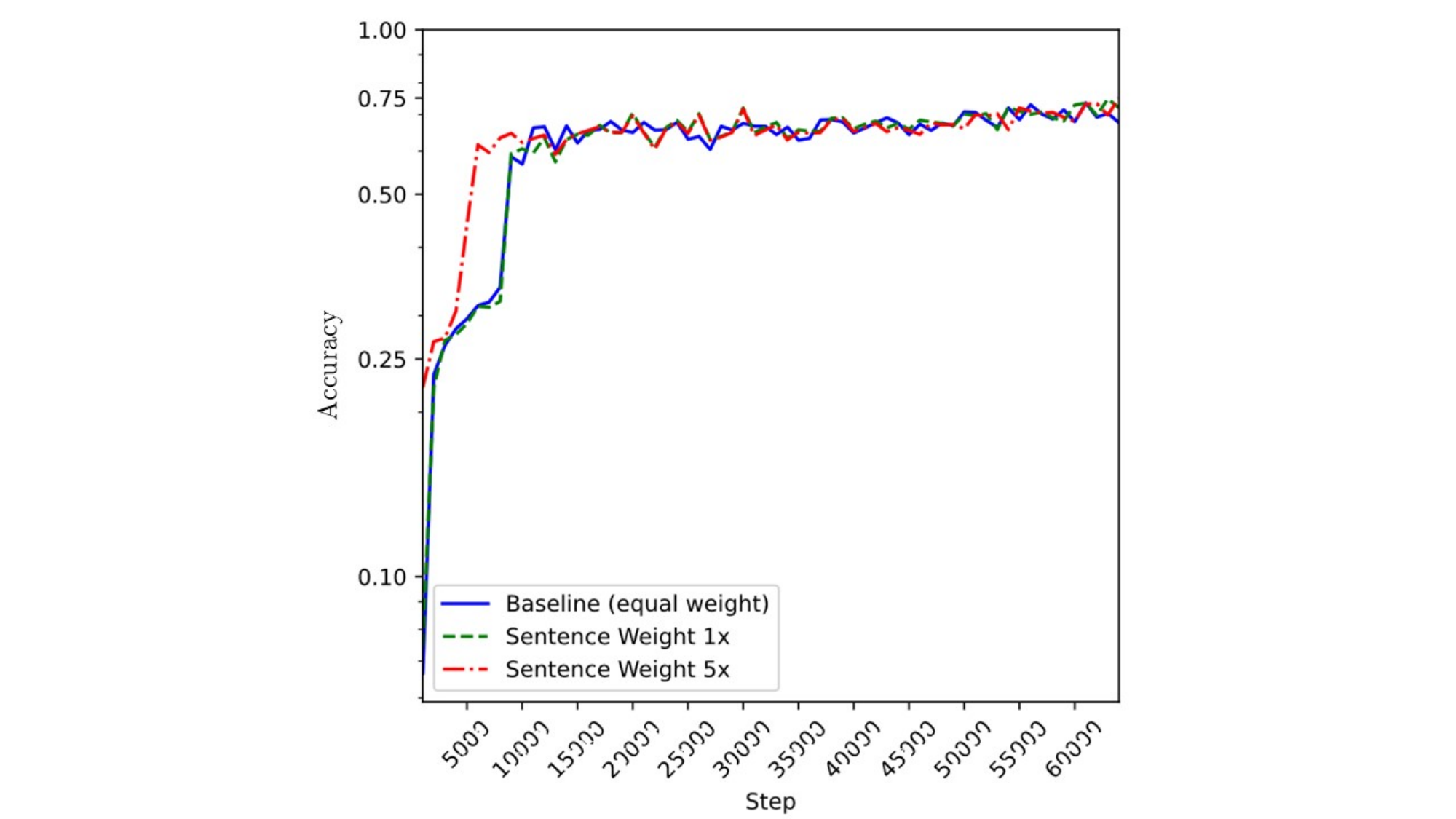}%
    }
    \caption{We explore the impact of weighing sentence loss more than reconstruction loss, and find that it has minimal impact beyond early stages of pre-training.}
    \label{fig:sentence_weight_experiment}
\end{figure}

We found that increasing the weight of this loss had minimal impact on the model's accuracy in predicting sentence tokens, and adversely harmed the model's loss (see Figure \ref{fig:sentence_weight_experiment}).

\section{Computational resources}
\label{sec:computational_resources}

We utilize 4 A40 GPUs, and 64 CPUs for training. We use a batch size of 200, since it helps us achieve much better GPU memory utilization. We leverage 16 CPUs for each GPU in order to increase the data loading time to accommodate for DEPTH's more-complex corruption method, and to allow effective optimizer offloading with DeepSpeed Zero2 \citep{zero_memory_optimization}.

\section{Pre-training considerations}
\label{sec:hyper_parameters}

\subsection{Pre-training scale}
\label{sec:pre_training_hyper_parameters}

In Table \ref{tab:pre_training_hparams} below we show the relative magnitude of DEPTH's pre-training. Specifically, we compare the number of observed tokens and optimization steps of our T5 and DEPTH models to comparably sized models (such as SLM, BERT, and RoBERTa). We highlight that the models we independently pre-trained (bottom 4 rows of the table) observed far fewer tokens than comparable LMs.

\begin{table*}[h!]
\vspace{1pt}
    \centering
    \begin{tabular}{lccccc}
    \toprule
    
     Model & Tokens & Steps & Batch Size & \# Params & Learning Rate \\
     
    \midrule
    
     SLM-1M & $125B$ & $1M$ & $256$ & $\approx 110M$ & $1.5e^{-4}$ \\
     
     SLM-3M & $375B$ & $3M$ & $256$ & $\approx 110M$ & $1.5e^{-4}$\\
     
     BERT-Base & $137B$ & $1M$ & $256$ & $\approx 110M$ & $1e^{-4}$\\
     
     BERT-Large & $137B$ & $1M$ & $256$ 
     & $\approx 340M$ & $1e^{-4}$ \\
     
     RoBERTa-Base & $2.2T$ & $500k$ & $8000$ & $\approx 110M$ & $1e^{-4}$ \\
     
     CONPONO (*) & - & $256k$ & $256$ & $\approx 110M$ & $1e^{-4}$ \\
     
     T5-Base & $1T$ & $1M$ & $2048$ & $\approx 220M$ & $1e^{-2}$ \\

     \midrule
     
     T5-FS & $58.6B$ & $1M$ & $200$ & $\approx 220M$ & $1e^{-4}$\\
     
     T5-CPT (*) & $15B$ & $256k$ & $200$ & $\approx 220M$ & $1e^{-4}$\\
     
     DEPTH-FS & $48.9B$ & $1M$ & $200$ & $\approx 220M$ & $1e^{-4}$\\
     
     DEPTH-CPT (*) & $12.5B$ & $256k$ & $200$ & $\approx 220M$ & $1e^{-4}$ \\
     
     \bottomrule
    \end{tabular}
    \caption{Hyper-parameters of comparable models to DEPTH. We show published hyper-parameters in the top rows of the table, and the models we train ourselves in the bottom of the table. We mark all models initialized from publically released pre-training models with (*). Note that CONPONO did not report the number of tokens it pre-trained on, so we exclude that value from the table above.}
    \label{tab:pre_training_hparams}
\end{table*}


\subsection{Pre-training data}
\label{sec:pre_training_data}

The rationale behind selecting C4 extends beyond its sheer volume and diversity. Given DEPTH's architectural roots in the T5 model, which was originally pre-trained on C4, leveraging the same dataset facilitates a direct comparison of the enhancements our model introduces. This baseline compatibility is crucial for isolating the effects of our architectural and methodological innovations on the model's performance. 
Furthermore, C4 is a subset of both Dolma \cite{dolma} and RedPajama \cite{together2023redpajama}. These datasets were used to train the most capable fully-open-source LMs (with released data) to date: OLMO \cite{Groeneveld2023OLMo} and LLama \cite{touvron2023llama} respectively. This suggests that C4 is an effective component in pre-training effective LMs, and offers multiple additional datasets components that would be new to our model for future training. Finally, \citet{muennighoff2023scaling} suggest that pre-training over the same data up to $4\times$ still improves model performance, while \citep{raffel2020exploring_t5}'s model has not completed even one full iteration over the C4 dataset. This hints at the viability of continuing to pre-train T5 on the same dataset it was already pre-trained on.

\section{Error Analysis}
\label{sec:error_analysis_appendix}

We select randomly select 30 examples from each task in DiscoEval, and manually inspect the nature of our models' errors. For example, in the SP task, we were interested in observing if a model tended to misunderstand clues from pronouns or transitions. For both DC and SP, we were also counted the number of examples where a human might find the LLM's answer reasonable (given the human-perceived coherence of the example).

\subsection{Sentence Position}

Consistent with the macro-level results from our DiscoEval fine-tuning experiments, DEPTH generally performs better in FS than T5. However, in the CPT setting, while DEPTH shows better reasoning in some cases, T5-CPT outperforms it, particularly on simpler tasks like SP-Wiki and SP-Rocstory, likely benefiting from more consistent pretraining. DEPTH's greater number of reasonable mistakes indicates its strength in engaging with complex discourse structures, but pronoun resolution and transition errors remain areas for further improvement. 

Given the 30 examples we've sampled, we categorize the types of errors our models tend to make. An error type is a reason by which a person might be able to infer the correct label. If the model predicts incorrectly given an ``obvious'' hint (e.g. introducing an entity that is referenced via a pronoun), then we categorize the error type based on that hint. In Table \ref{tab:sp_error_analysis} we show the counts of error types that each model made on each subset of SP.

\begin{table}[ht]
    \centering
    \resizebox{\columnwidth}{!}{%
    \begin{tabular}{l c c c c c c }
        \toprule
        \textbf{Dataset} & \textbf{Model} & \textbf{0} & \textbf{1} & \textbf{2} & \textbf{3} & \textbf{4} \\ 
        \midrule
        
        \multirow{2}{*}{SP-Wiki CPT} & T5    & 0.57 & 0.07 & 0.23 & 0.13 & 0.07 \\
        & DEPTH & 0.57 & 0.03 & 0.23 & 0.2 & 0.03 \\ 
        \midrule
        
        \multirow{2}{*}{SP-Wiki FS}  & T5    & 0.57 & 0.07 & 0.23 & 0.13 & 0.07 \\
        & DEPTH & 0.57 & 0.03 & 0.23 & 0.2 & 0.03 \\
        \midrule
        
        \multirow{2}{*}{SP-Arxiv CPT} & T5    & 0.63 & 0.07 & 0.2 & 0.1 & 0 \\ 
        & DEPTH & 0.57 & 0.13 & 0.2 & 0.1 & 0 \\
        \midrule
        
        \multirow{2}{*}{SP-Arxiv FS}  & T5    & 0.37 & 0.23 & 0.3 & 0.07 & 0.03 \\ 
        & DEPTH & 0.43 & 0.1 & 0.3 & 0.23 & 0 \\
        \midrule
        
        \multirow{2}{*}{SP-Rocstory CPT} & T5    & 0.77 & 0.07 & 0.1 & 0.07 & 0 \\
        & DEPTH & 0.57 & 0.13 & 0.2 & 0.1 & 0 \\
        \midrule
        
        \multirow{2}{*}{SP-Rocstory FS}  & T5    & 0.47 & 0.2 & 0.27 & 0.07 & 0 \\
        & DEPTH & 0.6 & 0.2 & 0.13 & 0.07 & 0 \\
        \bottomrule

    \end{tabular}
    }
    \caption{Approximate ratio of prediction types for SP-Wiki, SP-Arxiv, and SP-Rocstory across T5 and DEPTH models in FS and CPT settings. There are 30 total examples. Each prediction over these examples is categorized into one of 5 prediction types: 0 - Correct prediction, 1 - Incorrect (hint from transitions), 2 - Incorrect (hint from pronoun), 3 - Incorrect (reasonable error), 4 - Incorrect (hint from punctuation).}
    \label{tab:sp_error_analysis}
\end{table}

\subsection{Discourse Coherence}

We found that in the Discourse Coherence (DC) subset, both models were strictly incorrect by predicting poorly formatted outputs. Each of the poorly formatted predictions was incorrect (e.g., if the model predicted ``cooherent'' instead of ``coherent'', the correct label was ``incoherent''). We analyze the ratio of these types of predictions on DC in Table \ref{tab:dc_prediction_types}. Further, we find that in DC-Chat, both models tend to make errors that a human might find reasonable. This implies that the augmentation on the input example (i.e., replacing one if the sentences in the paragraph with another one) did not adversely impact the example's coherence. Strangely, we found that FS models predicted more correct outputs than their CPT counterparts. This may be the result of selecting a too small a sample-size of examples to analyze.

\begin{table}[h!]
    \centering
    \resizebox{\columnwidth}{!}{%
    \begin{tabular}{l l c c c}
        \toprule
        \textbf{Dataset} & \textbf{Model} & \textbf{0} & \textbf{1} & \textbf{2} \\ \midrule
        
        \multirow{2}{*}{DC-Wiki-FS}  & T5     & 0.87 & 0.32 & 0.00 \\
        & DEPTH  & 0.87 & 0.13 & 0.00 \\ \midrule
        
        \multirow{2}{*}{DC-Wiki-CPT} & T5     & 0.80 & 0.08 & 0.00 \\
        & DEPTH  & 0.83 & 0.08 & 0.00 \\ \midrule
        
        \multirow{2}{*}{DC-Chat-FS}  & T5     & 0.60 & 0.42 & 0.17 \\
        & DEPTH  & 0.63 & 0.35 & 0.23 \\ \midrule
        
        \multirow{2}{*}{DC-Chat-CPT} & T5     & 0.83 & 0.30 & 0.10 \\
        & DEPTH  & 0.77 & 0.29 & 0.07 \\ 
        
        \bottomrule
    \end{tabular}
    }
    \caption{Prediction types for models and dataset splits. 0: Correct predictions (Type 0), 1: Poorly formatted predictions (Type 1), 2: Reasonable errors (Type 2).}
    \label{tab:dc_prediction_types}
\end{table}

\section{GLUE results}

In this section we show the full results from our downstream experiments on GLUE tasks. Table \ref{tab:glue_results} is reflected in the top row (FS) of Figure \ref{fig:combined_glue_results}, while Table \ref{tab:glue_cpt_results} is reflected in the bottom row (CPT) of figure \ref{fig:combined_glue_results}. Consistent with our hypothesis, we found that both DEPTH and T5 improve across downstream tasks as a function of the pre-training steps they've taken. While DEPTH outperforms T5 across all tasks is the FS setting, it did not reach the scores of \citet{raffel2020exploring_t5}'s T5 model. In the CPT setting, T5 and DEPTH perform quite comparably. In fact, in the penultimate checkpoint (128k) we found that DEPTH outperformed T5 on all tasks except for CoLA.

\label{sec:glue_results_appendix}

\begin{table*}[ht!]
\centering
\begin{tabularx}{\textwidth}{l *{4}{>{\centering\arraybackslash}X}}
\toprule

\textbf{Model} & \textbf{CoLA} & \textbf{SST-2} & \multicolumn{2}{c}{\textbf{MNLI}} \\

& & & \textit{Matched} & \textit{Mismatched} \\ 

\midrule

T5-Base @ 0 (Test) & 12.3 & 80.62 & 68.02 & 68.0 \\


T5-Base @ 1M (Test) & 53.84 & 92.68 & 84.24 & 84.57 \\

T5-Base @ 1M (Val) & 53.98 & 94.73 & 87.28 & 87.1 \\

\midrule

T5 @ 2k & 8.62 & 80.08 & 52.54 & 52.3 \\

DEPTH @ 2k & 8.67 & 78.52 & 57.07 & 58.14 \\

\midrule

T5 @ 4k & 6.23 & 80.66 & 53.16 & 53.89 \\

DEPTH @ 4k & 7.75 & 80.08 & 59.31 & 59.62 \\

\midrule

T5 @ 8k & 4.66 & 82.23 & 54.33 & 54.62 \\

DEPTH @ 8k & 6.93 & 79.69 & 58.92 & 59.73 \\

\midrule

T5 @ 16k & 8.99 & 80.96 & 54.1 & 54.1 \\

DEPTH @ 16k & 10.94 & 81.25 & 62.79 & 61.46 \\

\midrule

T5 @ 32k & 10.72 & 81.64 & 55.18 & 55.6 \\

DEPTH @ 32k & 7.73 & 82.81 & 71.23 & 72.7 \\

\midrule

T5 @ 64k & 6.86 & 82.42 & 57.68 & 60.88 \\

DEPTH @ 64k & 27.78 & 86.72 & 73.84 & 76.05 \\

\midrule

T5 @ 128k & 12.85 & 83.2 & 69.61 & 69.82 \\

DEPTH @ 128k & 38.01 & 88.87 & 77.5 & 78.06 \\

\midrule

T5 @ 256k & 11.78 & 85.94 & 72.82 & 73.39 \\

DEPTH @ 256k & 45.57 & 91.31 & 79.45 & 80.07 \\

\midrule

T5 @ 512k & 19.96 & 86.52 & 74.22 & 74.26 \\

DEPTH @ 512k & 47.14 & 91.11 & 80.42 & 81.43 \\

\midrule

T5 @ 1M & 29.35 & 88.77 & 74.53 & 75.37 \\

DEPTH @ 1M & 45.91 & 91.41 & 81.0 & 81.96 \\

\bottomrule
\end{tabularx}
\caption{GLUE benchmark results for From Scratch (FS). Note that the first two rows are reported by Raffel et al., 2019, while all later rows are the best reported results on the validation set across 3 attempted learning rates.}
\label{tab:glue_results}
\end{table*}

\begin{table*}[h!]
\centering
\begin{tabularx}{\textwidth}{l *{4}{>{\centering\arraybackslash}X}}
\toprule

\textbf{Model} & \textbf{CoLA} & \textbf{SST-2} & \multicolumn{2}{c}{\textbf{MNLI}} \\

& & & \textit{Matched} & \textit{Mismatched} \\ 

\midrule

T5-Base @ 1M (Test) & 53.84 & 92.68 & 84.24 & 84.57 \\

T5-Base @ 1M (Val) & 53.98 & 94.73 & 87.28 & 87.1 \\

\midrule

T5 @ 2k & 57.41 & 95.02 & 86.93 & 87.06 \\

DEPTH @ 2k & 53.93 & 94.34 & 86.38 & 86.2 \\

\midrule

T5 @ 4k & 55.18 & 95.02 & 87.4 & 87.34 \\

DEPTH @ 4k & 47.11 & 94.34 & 87.14 & 87.01 \\

\midrule

T5 @ 8k & 55.35 & 95.21 & 87.47 & 87.31 \\

DEPTH @ 8k & 50.67 & 94.43 & 86.62 & 86.52 \\

\midrule

T5 @ 16k & 54.75 & 95.7 & 86.91 & 86.67 \\

DEPTH @ 16k & 52.65 & 94.34 & 86.62 & 86.67 \\

\midrule

T5 @ 32k & 54.95 & 95.21 & 86.64 & 86.06 \\

DEPTH @ 32k & 53.79 & 94.63 & 86.95 & 87.02 \\

\midrule

T5 @ 64k & 54.77 & 95.21 & 86.96 & 86.93 \\

DEPTH @ 64k & 52.95 & 94.14 & 86.65 & 86.91 \\

\midrule

T5 @ 128k & 58.62 & 95.21 & 86.79 & 86.67 \\

DEPTH @ 128k & 56.21 & 95.61 & 87.42 & 87.64 \\

\midrule

T5 @ 256k & 57.62 & 95.21 & 87.27 & 87.22 \\

DEPTH @ 256k & 56.78 & 95.02 & 86.86 & 86.45 \\

\bottomrule
\end{tabularx}
\caption{GLUE benchmark results for Continuous Pre-Training (CPT). As in the FS setting, we report our results on the validation set after a hyper-parameter sweep over 3 learning rates.}
\label{tab:glue_cpt_results}
\end{table*}

\section{DiscoEval results}
\label{sec:discoeval_results_appendix}

In this section we show the full results from our downstream experiments on discourse tasks from the DiscoEval benchmark. In Table \ref{tab:discoeval_fs_results}, we show the full results of our models in the FS setting, while in Table \ref{tab:discoeval_cpt_results} we show the full results of our models in the CPT setting. These tables reflect the top and bottom row of Figure \ref{fig:combined_discoeval_results} respectively. 

In the FS setting, we found that DEPTH's wins over T5 are even more pronounced in DiscoEval than they were in GLUE. Specifically, T5 struggles to learn discourse tasks (especially SP) during early stages of pre-training. On the other hand, DEPTH was highly effective in discourse tasks already from early pre-training checkpoints. In the CPT setting, we found that DEPTH still outperformed T5, despite the fact that the original checkpoint was pre-trained substantially with a different objective. 

\begin{table*}[htbp]
\centering
\begin{tabularx}{\textwidth}{l *{5}{>{\centering\arraybackslash}X}}
\toprule
\textbf{Model} & \multicolumn{3}{c}{\textbf{SP}} & \multicolumn{2}{c}{\textbf{DC}} \\
  & \textit{Arxiv} & \textit{Wiki} & \textit{Rocstory} & \textit{Chat} & \textit{Wiki} \\
\midrule

Baseline T5 @ 1M & 52.76 & 51.07 & 70.58 & 68.99 & 92.09 \\

\midrule

T5 @ 2k & 21.0 & 20.9 & 21.2 & 55.18 & 53.59 \\

DEPTH @ 2k & 34.81 & 40.19 & 51.51 & 55.27 & 53.2 \\

\midrule

T5 @ 4k & 20.4 & 21.4 & 20.9 & 57.18 & 55.37 \\

DEPTH @ 4k & 35.72 & 40.38 & 52.03 & 56.49 & 54.74 \\

\midrule
T5 @ 8k & 20.6 & 20.9 & 21.0 & 57.42 & 54.57 \\

DEPTH @ 8k & 36.43 & 40.14 & 51.25 & 57.03 & 54.69 \\

\midrule
T5 @ 16k & 21.4 & 22.0 & 21.24 & 57.62 & 55.18 \\

DEPTH @ 16k & 36.47 & 40.33 & 53.13 & 57.52 & 55.44 \\

\midrule

T5 @ 32k & 21.75 & 21.75 & 20.63 & 57.62 & 56.1 \\

DEPTH @ 32k & 38.04 & 44.26 & 54.05 & 58.5 & 57.37 \\

\midrule

T5 @ 64k & 21.92 & 21.53 & 21.24 & 57.23 & 57.01 \\

DEPTH @ 64k & 42.24 & 45.85 & 55.96 & 60.94 & 72.58 \\

\midrule

T5 @ 128k & 21.09 & 33.96 & 42.63 & 60.16 & 60.01 \\

DEPTH @ 128k & 45.0 & 47.71 & 59.57 & 64.65 & 78.0 \\

\midrule

T5 @ 256k & 22.85 & 37.92 & 44.85 & 58.89 & 61.91 \\

DEPTH @ 256k & 48.68 & 48.93 & 61.33 & 65.33 & 81.69 \\

\midrule

T5 @ 512k & 26.61 & 41.97 & 52.29 & 61.33 & 60.89 \\

DEPTH @ 512k & 52.59 & 51.66 & 65.92 & 66.85 & 83.81 \\

\midrule

T5 @ 1M & 28.54 & 43.22 & 50.98 & 61.13 & 65.48 \\

DEPTH @ 1M & 52.39 & 50.07 & 63.89 & 67.92 & 84.52 \\

\hline
\end{tabularx}
\caption{DiscoEval Downstream Full Training (FS) Results}
\label{tab:discoeval_fs_results}
\end{table*}

\begin{table*}[htbp]
\centering
\begin{tabularx}{\textwidth}{l *{5}{>{\centering\arraybackslash}X}}
\toprule
\textbf{Model} & \multicolumn{3}{c}{\textbf{SP}} & \multicolumn{2}{c}{\textbf{DC}} \\
& \textbf{Arxiv} & \textbf{Wiki} & \textbf{Rocstory} & \textbf{Wiki} & \textbf{Chat} \\
\hline

T5 @ 2k & 62.26 & 51.9 & 74.83 & 92.09 & 71.44 \\

DEPTH @ 2k & 44.14 & 49.34 & 74.78 & 89.99 & 70.46 \\

\midrule

T5 @ 4k & 61.33 & 52.39 & 74.98 & 91.72 & 74.27 \\

DEPTH @ 4k & 56.62 & 51.2 & 67.65 & 91.46 & 72.02 \\

\midrule

T5 @ 8k & 63.63 & 52.03 & 77.0 & 92.33 & 73.1 \\

DEPTH @ 8k & 55.03 & 51.56 & 71.02 & 91.99 & 71.97 \\

\midrule

T5 @ 16k & 60.94 & 51.78 & 76.49 & 91.53 & 74.71 \\

DEPTH @ 16k & 58.86 & 52.08 & 76.15 & 92.63 & 72.71 \\

\midrule

T5 @ 32k & 60.69 & 52.22 & 75.61 & 92.33 & 74.02 \\

DEPTH @ 32k & 59.35 & 53.27 & 78.08 & 92.48 & 72.85 \\

\midrule

T5 @ 64k & 58.96 & 52.25 & 76.07 & 92.58 & 73.73 \\

DEPTH @ 64k & 58.57 & 52.95 & 76.1 & 92.58 & 73.24 \\

\midrule

T5 @ 128k & 60.13 & 51.03 & 75.76 & 91.14 & 73.44 \\

DEPTH @ 128k & 70.56 & 54.77 & 82.42 & 92.53 & 73.96 \\

\midrule

T5 @ 256k & 59.03 & 52.66 & 66.77 & 92.31 & 72.22 \\

DEPTH @ 256k & 67.07 & 53.0 & 76.71 & 92.07 & 72.9 \\

\bottomrule
\end{tabularx}
\caption{DiscoEval Downstream Continuous Pre-Training (CPT) Results}
\label{tab:discoeval_cpt_results}
\end{table*}

\newpage

\section{NI results}
\label{sec:ni_results_appendix}

In the from-scratch setting (Figure \ref{fig:ni_fs} and Table \ref{tab:ni_fs}), we observe that DEPTH outperforms T5 in the NI benchmark, with a notable leap in performance between steps 16k and 32k. 
This indicates that DEPTH's pre-training objective is more effective at learning representations that are beneficial for the NI task. However, at steps 2k, 8k, and 16k, DEPTH underperforms compared to T5, suggesting that the benefits of DEPTH's pre-training objective may not be immediately apparent in the early stages of training. 

However, in the continuously pre-trained setting (Figure \ref{fig:ni_cpt} and Table \ref{tab:ni_cpt}), we find that DEPTH's pre-training harms downstream performance compared to T5. 
Additionally, we observe that CPT models are less sensitive to learning rate and can train effectively across a wider range of learning rates, with the exception of DEPTH in the early stages of CPT, where it is adapting to a task that differs from its initial pre-training.
This robustness to learning rate is a positive property that the FS models did not exhibit, likely due to limitations in training scale (e.g., small batch size, avoiding packing, and training on fewer tokens overall). Furthermore, early in the CPT process, DEPTH's performance is somewhat unstable, possibly due to the domain shift from T5's pre-training task to DEPTH's pre-training task. Interestingly, lower learning rates perform worse for DEPTH after CPT, suggesting that the model needs to adjust its representations more substantially to adapt to the downstream task. 

\begin{table}[ht!]
\centering
\begin{tabularx}{0.4\textwidth}{l c >{\centering\arraybackslash}X}
\toprule
\textbf{Model} & \textbf{Step} & \textbf{RougeL} \\
\midrule
Baseline T5 & 1M & 42.48 \\
\midrule
T5 & 2k & 8.36 \\
DEPTH & 2k & 10.92 \\
\midrule
T5 & 4k & 10.11 \\
DEPTH & 4k & 11.03 \\
\midrule
T5 & 8k & 10.15 \\
DEPTH & 8k & 9.06 \\
\midrule
T5 & 16k & 10.82 \\
DEPTH & 16k & 10.43 \\
\midrule
T5 & 32k & 10.68 \\
DEPTH & 32k & 23.51 \\
\midrule
T5 & 64k & 12.89 \\
DEPTH & 64k & 30.23 \\
\midrule
T5 & 128k & 18.24 \\
DEPTH & 128k & 32.24 \\
\midrule
T5 & 256k & 26.36 \\
DEPTH & 256k & 32.63 \\
\midrule
T5 & 512k & 28.05 \\
DEPTH & 512k & 34.72 \\
\midrule
T5 & 1M & 29.6 \\
DEPTH & 1M & 33.8 \\
\bottomrule
\end{tabularx}
\caption{NI benchmark results for From Scratch (FS) pre-training. The first row reports the performance of the baseline T5 model, while all later rows show the best reported results on the validation set across 3 attempted learning rates.}
\label{tab:ni_fs}
\end{table}

\begin{table}[ht!]
\centering
\begin{tabularx}{0.4\textwidth}{l c >{\centering\arraybackslash}X}
\toprule
\textbf{Model} & \textbf{Step} & \textbf{NI RougeL}\\
\midrule
T5 & 2k & 41.96\\
DEPTH & 2k & 39.15\\
\midrule
T5 & 4k & 42.72\\
DEPTH & 4k & 37.85\\
\midrule
T5 & 8k & 42.83\\
DEPTH & 8k & 38.04\\
\midrule
T5 & 16k & 42.88\\
DEPTH & 16k & 38.19\\
\midrule
T5 & 32k & 43.56\\
DEPTH & 32k & 37.79\\
\midrule
T5 & 64k & 43.06\\
DEPTH & 64k & 38.99\\
\midrule
T5 & 128k & 42.58\\
DEPTH & 128k & 39.19\\
\midrule
T5 & 256k & 43.29\\
DEPTH & 256k & 37.86\\
\bottomrule
\end{tabularx}
\caption{NI benchmark results for Continuous Pre-Training (CPT). All rows show the best reported results on the validation set across 3 attempted learning rates.}
\label{tab:ni_cpt}
\end{table}

\end{document}